\documentclass[journal]{IEEEtran}

\usepackage{cite}
\usepackage{url}
\usepackage[hidelinks]{hyperref}
\usepackage{xcolor}
\usepackage{amssymb}
\usepackage{graphicx}
\usepackage{amsmath}
\usepackage{multirow}
\usepackage{enumerate}
\usepackage{pifont}
\usepackage{bbm}
\usepackage{booktabs}
\usepackage{floatrow}
\usepackage{efbox}
\efboxsetup{linecolor=blue,linewidth=2pt}

\graphicspath{{./images/}}

\usepackage{array}
\newcommand{\PreserveBackslash}[1]{\let\temp=\\#1\let\\=\temp}
\newcolumntype{C}[1]{>{\PreserveBackslash\centering}p{#1}}
\newcolumntype{R}[1]{>{\PreserveBackslash\raggedleft}p{#1}}
\newcolumntype{L}[1]{>{\PreserveBackslash\raggedright}p{#1}}

\hypersetup{
    colorlinks=true,%
    urlcolor=magenta
}

\begin{document}

\title{Instance-level Heterogeneous Domain Adaptation for Limited-labeled Sketch-to-Photo Retrieval}

\author{Fan~Yang,~\IEEEmembership{Member,~IEEE,}
        Yang~Wu,~\IEEEmembership{Member,~IEEE,} 
        Zheng~Wang,~\IEEEmembership{Member,~IEEE,}
        Xiang~Li, \\
        Sakriani~Sakti,~\IEEEmembership{Member,~IEEE,}
        and Satoshi~Nakamura,~\IEEEmembership{Fellow,~IEEE}

}

\markboth{IEEE}%
{Yang \MakeLowercase{\textit{et al.}}: Instance-level Heterogeneous Domain Adaptation}

\maketitle

\begin{abstract}
Although sketch-to-photo retrieval has a wide range of applications, it is costly to obtain paired and rich-labeled ground truth. Differently, photo retrieval data is easier to acquire. Therefore, previous works pre-train their models on rich-labeled photo retrieval data (i.e., source domain) and then fine-tune them on the limited-labeled sketch-to-photo retrieval data (i.e., target domain). However, without co-training source and target data, source domain knowledge might be forgotten during the fine-tuning process, while simply co-training them may cause negative transfer due to domain gaps. Moreover, identity label spaces of source data and target data are generally disjoint and therefore conventional category-level Domain Adaptation (DA) is not directly applicable. To address these issues, we propose an Instance-level Heterogeneous Domain Adaptation (IHDA) framework. We apply the fine-tuning strategy for identity label learning, aiming to transfer the instance-level knowledge in an inductive transfer manner. Meanwhile, labeled attributes from the source data are selected to form a shared label space for source and target domains. Guided by shared attributes, DA is utilized to bridge cross-dataset domain gaps and heterogeneous domain gaps, which transfers instance-level knowledge in a transductive transfer manner. Experiments show that our method has set a new state of the art on three sketch-to-photo image retrieval benchmarks without extra annotations, which opens the door to train more effective models on limited-labeled heterogeneous image retrieval tasks. \emph{Related codes are available at \url{https://github.com/fandulu/IHDA}.}
\end{abstract}


\section{Introduction}
\label{sec:intro}
\begin{figure}[t]
\centering
  \includegraphics[width=\columnwidth]{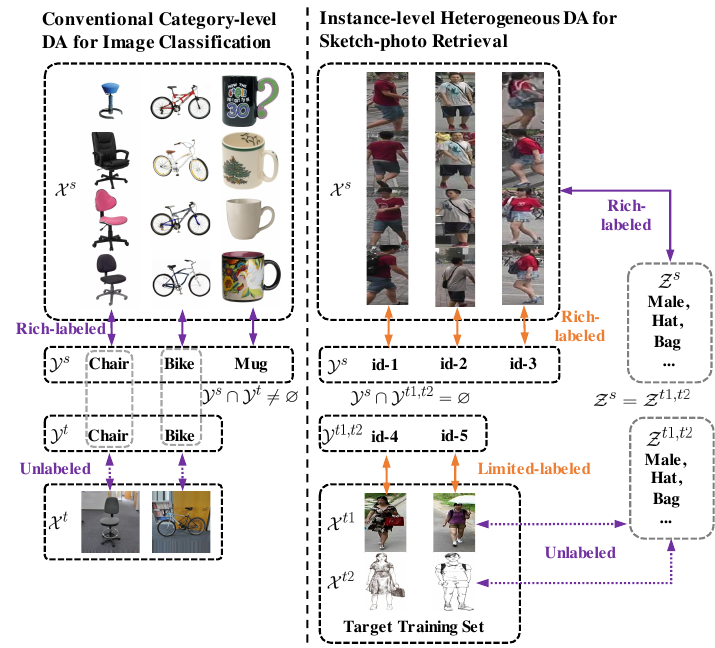}
  \caption{Comparison of Category-level DA (left) and IHDA (right). The solid two-way arrow connector and the dotted two-way arrow connector represent labeled and unlabeled, resp. The \textcolor{orange}{orange} two-way arrow connector and the \textcolor{violet}{purple} two-way arrow connector represent inductive transfer flow and transductive transfer flow, resp. }
  \label{fig:demo}
\end{figure}

Domain Adaptation (DA)~\cite{pan2010survey} is used to transfer the knowledge of the rich-labeled source domain ($\mathcal{D}^{s}$) to a unlabeled/limited-labeled target domain ($\mathcal{D}^{t}$) and improve the effectiveness of a given task in the target domain. Conventional DA studies mainly focus on the category-level cases, such as image classification and segmentation tasks~\cite{pei2018multi,rozantsev2018beyond,cao2018partial,zou2019consensus,zou2018unsupervised}. In category-level DA, the target label space ($\mathcal{Y}^{t}$) and the source label space ($\mathcal{Y}^{s}$) are either fully shared or partially shared, which can be represented as $\mathcal{Y}^{s} \cap \mathcal{Y}^{t} \!\neq\! \varnothing$ (see Fig.~\ref {fig:demo} and the notations used in this section are summarized in TABLE \ref{tab:notations}). 

In real-world applications, the needs of DA are not limited to category-level tasks. For instance, the instance-level sketch-to-photo retrieval, which has been investigated in many studies~\cite{pang2018cross,ouyang2016forgetmenot,wang2015sketch,yu2016sketch,zhang2016sketch,bhattacharjee2018query,choi2019sketchhelper,wang2020beyond}, is suffering lack of training data as it is costly to draw sketches. Meanwhile, plenty of rich-labeled photo retrieval datasets are available. In many existing photo retrieval datasets~\cite{semjitter,liu2015faceattributes,zheng2015scalable,wang2019incremental,zeng2020illumination}, even instance attributes are annotated~\cite{wang2015multi,wang2018incremental}. Accordingly, majority of sketch-to-photo retrieval methods~\cite{yu2016sketch,song2016deep,song2017deep,pang2018cross,wu2018light,wu2018coupled,deng2019residual} choose to pre-train their models on a rich-labeled photo retrieval data (i.e., source domain) and then fine-tune them on the limited-labeled sketch-to-photo retrieval data (i.e., target domain). However, without co-training source and target data, source domain knowledge might be forgotten during the fine-tuning process, while simply co-training them may cause negative transfer due to domain gaps.

DA methods are commonly used to reduce domain gaps on co-training source and target datasets. sketch-to-photo retrieval data consists of two heterogeneous modalities as sketch and photo. Accordingly, we divide the target data into two sub-domains as $\mathcal{D}^{t1}$ (i.e., photos) and $\mathcal{D}^{t2}$ (i.e., sketches), where $\mathcal{D}^t=\mathcal{D}^{t1} \cup \mathcal{D}^{t2}$. The feature spaces of $\mathcal{D}^{s}$, $\mathcal{D}^{t1}$, and $\mathcal{D}^{t2}$ are $\mathcal{X}^{s}$, $\mathcal{X}^{t1}$ and $\mathcal{X}^{t2}$, respectively. If we regard each identity as one class, the identity label space of $\mathcal{X}^{s}$ is $\mathcal{Y}^{s}$ while $\mathcal{X}^{t1}$ and $\mathcal{X}^{t2}$ share the same identity label space $\mathcal{Y}^{t1,t2}$. Nonetheless, the source data is annotated by other contributors and identity label spaces of the source and target domains are generally disjoint (i.e., $\mathcal{Y}^{s} \cap \mathcal{Y}^{t1,t2} = \varnothing$). It is challenging for category-level DA methods to figure out ``what to transfer and where to transfer'' in this case. It shows that coarsely reducing domain gaps may incur a target error~\cite{saito2017asymmetric}. It is insufficient to simply match the marginal distributions, conditional distributions of both domains should also be matched.

To address this issue, we raise an Instance-level Heterogeneous Domain Adaptation (IHDA) framework, which breaks through limitations of simple fine-tuning and category-level DA. We select attributes that can jointly characterize instance-level properties for all domains to form a shared label space as $\mathcal{Z}^{s}=\mathcal{Z}^{t1,t2}$, where $\mathcal{Z}^{s}$ and $\mathcal{Z}^{t1,t2}$ are proposed attribute spaces for source domain and target domain, respectively. For instance-level retrieval problems, since all of the instances belong to the same category, the existence of shared attributes is guaranteed. We assume that the attributes of the source domain are observed. For some photo-sketch retrieval applications, a lot of related photo retrieval datasets already offer attributes~\cite{liu2015faceattributes,lin2019improving}. When attributes of the source dataset are not available, it is still possible to collect it from large web resources at an acceptable cost. As many objects are accompanied by attributes on website resources, both photos and attributes can be downloaded to construct the source data. For instance, the UT-Zap50K dataset~\cite{semjitter}, a source data in our experiment, was collected from a shopping website.

After $\mathcal{Z}^{s}$ and $\mathcal{Z}^{t1,t2}$ are formed, our IHDA framework seeks to maximally transfer domain knowledge from source data to target data. Due to $\mathcal{Y}^{s} \cap \mathcal{Y}^{t1,t2} = \varnothing$, $\mathcal{Y}^{s}$ and  $\mathcal{Y}^{t1,t2}$ are distinct but related. Learning to identify $\mathcal{Y}^{s}$ is inductive to identify $\mathcal{Y}^{t1,t2}$. Thus, pre-training identify labels on $\mathcal{Y}^{s}$ and then fine-tuning identity labels on $\mathcal{Y}^{t1,t2}$ can transfer the instance-level knowledge from an inductive transfer manner. Since $\mathcal{Z}^{s}=\mathcal{Z}^{t1,t2}$, referring to them, cross-dataset domain gaps (i.e., $\mathcal{D}^{s}$ to $\mathcal{D}^{t1}$ and $\mathcal{D}^{t2}$) and heterogeneous domain gaps (i.e., $\mathcal{D}^{s}$ and $\mathcal{D}^{t1}$ to $\mathcal{D}^{t2}$) can be reduced via an unsupervised Domain Adaptation~\cite{pei2018multi,rozantsev2018beyond,cao2018partial,zou2019consensus}, which transfers instance-level knowledge from a transductive transfer manner.

Although the ground truth of $\mathcal{Z}^{t1,t2}$ is unknown, it is leveraged to associate source and target domains by using an entropy minimization criterion~\cite{luo2017label}. The entropy minimization criterion does penalize low-confident predictions of $\mathcal{Z}^{t1,t2}$. In order to minimize it, target domain features are forced to match source domain features at an instance level other than roughly reducing the cross-domain gaps. Such an approach remarkably improves the IHDA performance. 

In summary, our work has \textbf{three contributions}:

\begin{itemize}
\item We highlight the challenges for improving limited-labeled sketch-to-photo retrieval performance from the transfer learning perspective.
\item We propose a novel IHDA framework to maximally utilize the source domain knowledge to benefit the target task at instance-level image retrieval. It overcomes the limitations of applying simple fine-tuning and conventional DA.
\item Our IHDA framework contributes new state-of-the-art results on instance-level sketch-to-photo retrieval task. It opens the door to train more effective cross-modal image retrieval models by using related rich-labeled single-modal image data. 
\end{itemize}

\tabcolsep=4pt
\begin{table}[!ht]
\centering
\caption{Summary of Notations.}
\label{tab:notations}
\resizebox{\linewidth}{!}{
\begin{tabular}{c|c}
\toprule
\textbf{Symbol} & \textbf{Description} \\ \midrule
$\mathcal{D}^{s}$ & Rich-labeled source domain\\
$\mathcal{D}^{t}$ & Limited-labelled target domain, $\mathcal{D}^t=\mathcal{D}^{t1} \cup \mathcal{D}^{t2}$\\
$\mathcal{D}^{t1}$ & Target photo domain, $\mathcal{D}^{t1} \in \mathcal{D}^{t}$\\
$\mathcal{D}^{t2}$ & Target sketch domain, $\mathcal{D}^{t2} \in \mathcal{D}^{t}$ \\
$\mathcal{X}^{t}$ & Target data space, $\mathcal{X}^t=\mathcal{X}^{t1} \cup \mathcal{X}^{t2}$\\
$\mathcal{X}^{t1}$ & Target photo data space, $\mathcal{X}^{t1} \in \mathcal{X}^{t}$\\
$\mathcal{X}^{t2}$ & Target sketch data space, $\mathcal{X}^{t2} \in \mathcal{X}^{t}$ \\
$\mathcal{Y}^{s}$ & Source identity label space\\
$\mathcal{Y}^{t1,t2}$ & Target photo-sketch identity label space, which is shared for $\mathcal{D}^{t1}$ and $\mathcal{D}^{t2}$\\
$\mathcal{Z}^{s}$ & Source attribute label space, which is shared for $\mathcal{D}^{s}$, $\mathcal{D}^{t1}$, and $\mathcal{D}^{t2}$\\
$\mathcal{Z}^{t1,t2}$ & Target photo-sketch attribute label space, which is shared for $\mathcal{D}^{s}$, $\mathcal{D}^{t1}$, and $\mathcal{D}^{t2}$\\
\bottomrule
\end{tabular}}
\end{table}

The rest of this paper is organized as follows. In Section~\ref{sec:related_works}, a brief review of related works is given. In Section~\ref{sec:methodology}, we illustrate the details of the proposed IHDA framework. Section~\ref{sec:experiments} reports experimental results to prove the effectiveness of our IHDA framework. In Section~\ref{sec:discusssion}, we discuss the application scope of our method. Finally, Section~\ref{sec:conclusion} concludes this paper.

\section{Related Works}
\label{sec:related_works}

We summarize related works and clarify their differences and relationships in TABLE~\ref{tab:related_workds}.
\tabcolsep=4pt
\begin{table}[!ht]
\centering
\caption{A comparison of related works.}
\label{tab:related_workds}
\resizebox{0.9\linewidth}{!}{
\begin{tabular}{c|c|c|c|c}
\toprule
\textbf{Methods} & \textbf{\#Modalities} & \begin{tabular}[c]{@{}c@{}}\textbf{DA \&}\\\textbf{\#Domains} \end{tabular}& \begin{tabular}[c]{@{}c@{}}\textbf{Target}\\\textbf{Attributes}\end{tabular}   & \begin{tabular}[c]{@{}c@{}}\textbf{Identities}  \\ \textbf{$\mathcal{Y}^{t} \cap  \mathcal{Y}^{s} = \varnothing$} \end{tabular} \\ \midrule
\cite{song2017deep}\cite{wu2018light}\cite{wu2018coupled}\cite{deng2019residual}   
&2 &w/o DA & - &-   \\\midrule
\cite{yu2016sketch}\cite{song2016deep}\cite{ouyang2016forgetmenot}\cite{liu2018deep}
&2 &w/o DA & Annotated &-   \\\midrule
\cite{pei2018multi}\cite{rozantsev2018beyond}\cite{cao2018partial}\cite{zou2019consensus} 
 &1, 2 &w/ DA, 2 & -  & No   \\\midrule
\cite{gebru2017fine}\cite{cui2018large}
&1 &w/ DA, 2 & Unknown &No \\\midrule
\begin{tabular}[c]{@{}c@{}}
\cite{Lin2018multi}\cite{wang2018transferable}
\cite{ge2020mutual}\cite{song2020unsupervised}
\end{tabular} 
&1 &w/ DA, 2 & Unknown  &Yes  \\\midrule
\textbf{Ours}  &\textbf{2} &\textbf{w/ DA, 3} & \textbf{Unknown} &\textbf{Yes} \\
\bottomrule
\end{tabular}}
\end{table}

\begin{figure*}[!htb]
    \centering
    \includegraphics[width=0.9\textwidth]{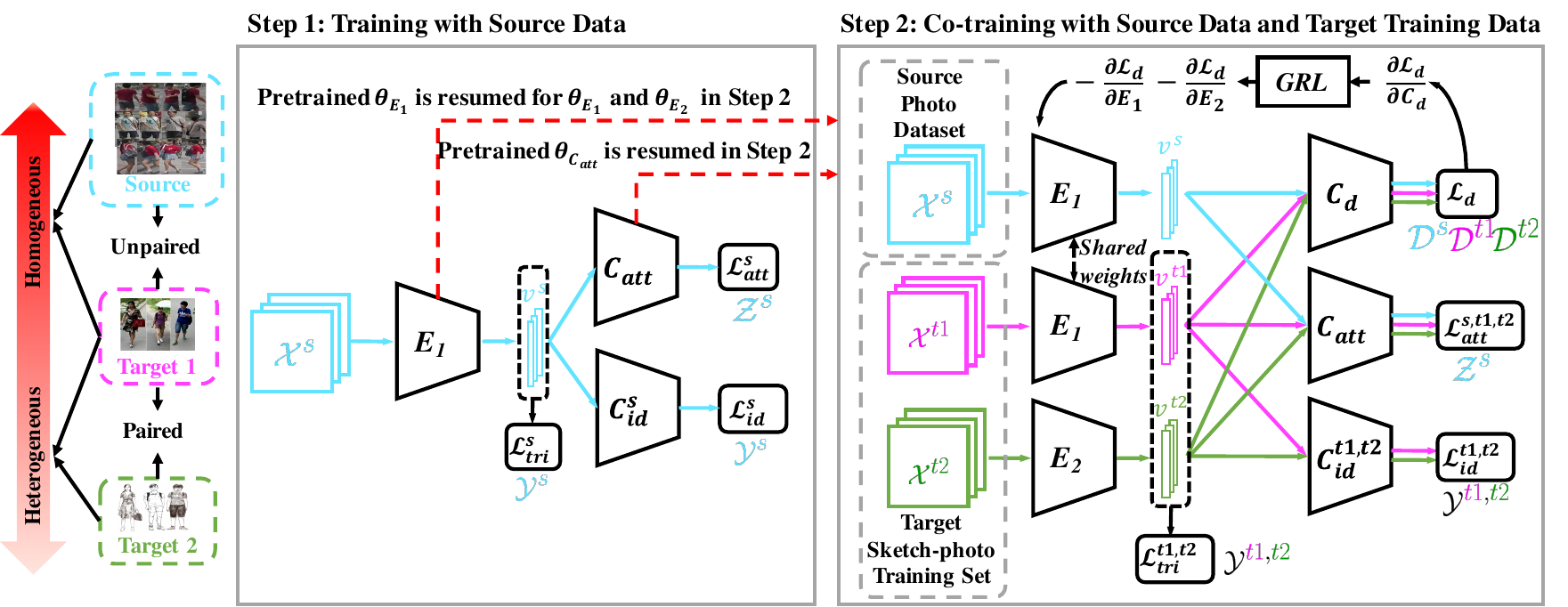}
 \caption{The framework of Instance-level Heterogeneous Domain Adaptation (IHDA). $E_{1}$ and $E_{2}$ are encoders for photos (i.e., $\mathcal{X}^{s}$ and $\mathcal{X}^{t1}$) and sketches (i.e., $\mathcal{X}^{t2}$), resp, and $v^{s}$, $v^{t1}$, and $v^{t2}$ denotes their embedding features; $C^{s}_{id}$ and $C^{t1,t2}_{id}$ stand for the identity classifier of source and target domains, resp; $C_{att}$ and $C_{d}$ denote the shared-attribute classifier and domain classifier, resp; $GRL$ is the Gradient Reversal Layer. Best viewed in color.}
  \label{fig:model}
\end{figure*}

Owing to the shortage of training data, studies~\cite{song2017deep,wu2018light,wu2018coupled,deng2019residual,yu2016sketch,song2016deep} take rich-labeled photo retrieval data for pre-training and then fine-tune their model on the sketch-to-photo retrieval data. However, there could be a dilemma in their approaches: performing a long-term fine-tuning may forget the source domain knowledge, which is known as catastrophic forgetting~\cite{lopez2017gradient,aljundi2018memory}, while a short-term fine-tuning may under-fit on the target data.

Alternatively, both the source and target data could be co-trained together. Due to the across-domain discrepancy, simply co-train both source and target data may lead to negative transfer~\cite{pan2010survey} and impair the performance. To alleviate this problem, Domain Adaptation (DA) is desired. Unlike coarse category-level DA~\cite{pei2018multi,rozantsev2018beyond,cao2018partial,zou2019consensus}, or fine-grained category-level DA~\cite{gebru2017fine,cui2018large}, applying DA in instance-level heterogeneous retrieval faces the challenge that $\mathcal{Y}^{s} \cap \mathcal{Y}^{t1,t2} = \varnothing$. To alleviate this challenge, some instance-level domain adaptation methods~\cite{ge2020mutual,song2020unsupervised} transferred the source domain knowledge to the target domain by using pseudo labels created by clustering algorithms on the target domain. However, clustering algorithms can easily generate noise in pseudo labels and impair the model performance. In our sketch-to-photo task, the noise label issue is more serious due to the heterogeneous domain gap. Therefore, instead of using pseudo labels, we specifically propose an IHDA framework, which selects annotated attributes in source data to form a shared label space for all domains.

Using attributes to learn the instance feature is particularly prominent in instance-level retrieval tasks. Before us, studies~\cite{yu2016sketch,song2016deep,ouyang2016forgetmenot,liu2018deep} had annotated attributes on both source and target data to improve retrieval performance. Compared with their methods, our IHDA is more practical since no labeled attribute is required for the target data, which mitigates the burden of manual labeling. Moreover, IHDA even outperforms their methods in our experiments. Previous works~\cite{Lin2018multi,wang2018transferable} also applied instance-level DA without annotated target attributes. However, their approaches focus on homogeneous DA scenario while our IHDA can tackle the more challenging instance-level heterogeneous DA problem. Besides, \cite{wang2018transferable} only adapts the marginal distribution while \cite{Lin2018multi} and ours jointly adapts the marginal distribution and conditional distribution, which can learn feature embedding that is robust for substantial distribution difference~\cite{long2013transfer,saito2019semi}.

We align conditional distribution by applying conditional entropy minimization~\cite{grandvalet2005semi,saito2019semi}, which can take the task-specific boundaries into account. IHDA forces source and target domain distributions to be matched at an instance level with the guidance of attributes. We further demonstrate such an approach does help to improve the model performance in our ablation studies.

\section{Methodology}
\label{sec:methodology}

We propose an Instance-level Heterogeneous Domain Adaptation (IHDA) framework as Fig.~\ref{fig:model} shows. Since $\mathcal{X}^{s}$ and $\mathcal{X}^{t1}$ are heterogeneous to $\mathcal{X}^{t2}$, we construct two separated encoder neural networks $E_{1}$ and $E_{2}$ respectively for $(\mathcal{X}^{s}, \mathcal{X}^{t1})$ and $\mathcal{X}^{t2}$. ResNet-50~\cite{he2016deep} is used as the backbone for both them and we denote their network parameters as $\theta_{E_{1}}$ and $\theta_{E_{2}}$. The inputs are represented as $x^{s}$, $x^{t1}$ and $x^{t2}$, where $x^{s} \in \mathcal{X}^{s}$, $x^{t1} \in \mathcal{X}^{t1}$ and $x^{t2} \in \mathcal{X}^{t2}$. IHDA takes two steps to transfer domain knowledge from the source domain to target domains. In the step 1, the encoder $E_{1}$, along with an identity classifier $C^{s}_{id}$ and an attribute classifier $C_{att}$, are pre-trained on the source data. In the step 2, the weights of the pre-trained attribute classifier are resumed, and the weights of $E_{1}$ are reloaded for both $E_{1}$ and $E_{2}$. The source and target training data are used to co-train the whole network by coupling an adversarial domain adaptation. We implement the adversarial domain adaptation by the Gradient Reversal Layer~\cite{ganin2015unsupervised}. \emph{Related codes are available at \url{https://github.com/fandulu/IHDA}.}

\subsection{Identity Learning}

We build identity classifiers $C^{s}_{id}$ and $C^{t1,t2}_{id}$ to perform identity classification. Each of them is a single dense layer, where the dimension of the output is equal to the number of identities. We denote the number of identities as $N^{s}_{id}$ and $N^{t1,t2}_{id}$ for the source and target training data, respectively. The network parameters of $C^{s}_{id}$ and $C^{t1,t2}_{id}$ are denoted as $\theta_{C^{s}_{id}}$ and $\theta_{C^{t1,t2}_{id}}$, respectively.

The identify classification losses $\mathcal{L}^{s}_{id}$ and $\mathcal{L}^{t1,t2}_{id}$ are defined as follows:
\begin{equation}
    \begin{split}
    \hat{\psi}^{s}& =  Softmax \bigg( C^{s}_{id} \big(E_{1}(x^{s}) \big) \bigg), \\
    \hat{\psi}^{t1}& =  Softmax \bigg(  C^{t1,t2}_{id} \big(E_{1}(x^{t1}) \big) \bigg), \\
\hat{\psi}^{t2} &=  Softmax \bigg(  C^{t1,t2}_{id} \big(E_{2}(x^{t2}) \big) \bigg), \\
    \mathcal{L}^{s}_{id} &=
         -\sum_{i=1}^{N^{s}_{id}} y^{s}_{i} \log (\hat{\psi}^{s}_{i}),\\
    \mathcal{L}^{t1,t2}_{id} &=
         -\sum_{i=1}^{N^{t1,t2}_{id}} y^{t1}_{i} \log (\hat{\psi}^{t1}_{i}) - \sum_{i=1}^{N^{t1,t2}_{id}} y^{t2}_{i} \log (\hat{\psi}^{t2}_{i}),\\
    \end{split}
\end{equation}
where $y^{s}$, $y^{t1}$ and $y^{t2}$ are one-hot ground truth labels of $x^s$, $x^{t1}$ and $x^{t2}$, respectively. The subscript $i$ indicates the $i_{th}$ element. $\hat{\psi}^{s}$, $\hat{\psi}^{t1}$ and $\hat{\psi}^{t2}$ are the corresponding predicted identity probabilities. Here, $y^{s}, \hat{\psi}^{s}\in \mathcal{Y}^{s}$ and $y^{t1}, y^{t2}, \hat{\psi}^{t1}, \hat{\psi}^{t2} \in \mathcal{Y}^{t1,t2}$.

In the step 1, $\mathcal{L}^{s}_{id}$ is used for the source data. In the step 2, since $\mathcal{Y}^{s} \cap \mathcal{Y}^{t1,t2} = \varnothing$, $\mathcal{L}^{t1,t2}_{id}$ is solely used for the target data. Apart from the identity classification, we also use a triplet loss to simultaneously learn joint embedding features that can better represent similarity and difference of identities. The joint embedding feature is a global average pooling feature with $2048$ dimensions. The triplet loss $\mathcal{L}^{s}_{tri}$, used in step 1, is the conventional triplet loss~\cite{chechik2010large} and we will not give its details here. 

We extend the conventional triplet loss to be a \emph{heterogeneous triplet loss} in step 2. For simplicity, let $v^{t1}\!=\!E_{1} (x^{t1})$ and $v^{t2}\!=\!E_{2}(x^{t2})$ stand for embedding features of photos and sketches, respectively. We represent the learned embedding feature space as $\mathcal{V}$, where $v^{t1},v^{t2} \in \mathcal{V}$. For feature-label pairs in a mini-batch of target training data, we sample a sketch anchor $(v^{t2}_a, y^{t2}_a)$, a photo positive sample $(v^{t1}_p, y^{t1}_p)$ and a photo negative sample $(v^{t1}_n, y^{t1}_n)$ with
\begin{equation}
\scalebox{1}{
\begin{math}
\begin{split}
y^{t1}_p &= y^{t2}_a ~\text{and}~ y^{t1}_n \neq y^{t2}_a, \\
\end{split}
\end{math}
}
\end{equation}
where $ y^{t2}_{a}, y^{t1}_{p},y^{t1}_{n} \in \mathcal{Y}^{t1,t2}$.

The heterogeneous triplet loss function is as follows,
\begin{equation}
\begin{split}
\mathcal{L}^{t1,t2}_{tri} &= \left [ \left \| v^{t2}_{a} -  v^{t1}_{p} \right \|_2^2 - \left \| v^{t2}_{a} -  v^{t1}_{n} \right \|_2^2 + \alpha \right ]_{+},\\
\end{split}
\end{equation}
where $\alpha$ indicates the margin between the positive and negative sketch-to-photo pairs. We empirically set $\alpha=0.3$.

\subsection{Attribute Learning}

Since selected attributes are shared among three domains, an attribute classifier $C_{att}$ is used for these domains in both step 1 and step 2. $C_{att}$ consists of three dense layers. For the first two layers, each has $512$ dimensions. The dimension of the output layer is $N_{att}$, which is the number of attributes. The network parameters of $C_{att}$ is represented as $\theta_{C_{att}}$.

Generally, each identity can be assigned to multiple attributes at the same time, we treat it as a multi-label classification task. Therefore, we apply a Binary Cross-Entropy loss for each kind of attribute and then sum them up as the multi-label classification loss. On the source domain, the attribute classification loss is 
\begin{equation}
\begin{split}
\hat{\phi}^{s} &=  Sigmoid \bigg(  C_{att} \big(E_{1}(x^{s}) \big) \bigg), \\
\mathcal{L}_{att}^{s} &= \sum_{i = 1}^{N_{att}}
-z_{i}^{s} \log (\hat{\phi}^{s}_{i}) - (1-z_{i}^{s}) \log ( 1- \hat{\phi}^{s}_{i}),\\
\end{split}
\end{equation}
where $z^{s}$ is one-hot ground truth label of $x^s$ and the subscript $i$ indicates the $i_{th}$ element; $\hat{\phi}^{s}$ is the estimated attributes. Here, $z^{s}, \hat{\phi}^{s} \in \mathcal{Z}^{s}$.

Without increasing annotations, attributes are unknown for the target data and therefore the loss function above cannot be applied directly. Inspired by previous works~\cite{luo2017label,shu2019transferable,zou2019consensus, saito2019semi}, a multi-label entropy minimization criterion is proposed to associate the attribute learning between the source domain and target domains. It does penalize low-confident predictions of target attributes. In order to minimize it, target domains are forced to match with the source domain at an instance level other than roughly reducing domain discrepancies, which leads to a better DA performance. On the target domains, the attribute classification loss is  
\begin{equation}
\begin{split}
\hat{\phi}^{t1} &=  Sigmoid \bigg(  C_{att} \big(E_{1}(x^{t1}) \big) \bigg), \\
\hat{\phi}^{t2} &=  Sigmoid \bigg(  C_{att} \big(E_{2}(x^{t2}) \big) \bigg), \\
\mathcal{L}_{att}^{t1,t2} &=  -\sum_{i = 1}^{N_{att}} \hat{\phi}^{t1}_{i} \log(\hat{\phi}^{t1}_{i}) -\sum_{i = 1}^{N_{att}} \hat{\phi}^{t2}_{i} \log(\hat{\phi}^{t2}_{i}), \\
\end{split}
\end{equation}
where $\hat{\phi}^{t1}$ and $\hat{\phi}^{t2}$ are estimated attributes of $x^{t1}$ and $x^{t2}$, respectively. Here, $\hat{\phi}^{t1},\hat{\phi}^{t2} \in \mathcal{Z}^{t1,t2}$.

Moreover, when the photo and sketch are paired in target data, their predicted attributes should be identical. Using this property, we further construct an attribute-consistent loss for paired target data as
\begin{equation}
\begin{split}
\mathcal{L}_{con}^{t1,t2} &=  \mathbbm{1}_{y^{t1}=y^{t2}} \left \| \hat{\phi}^{t1}- \hat{\phi}^{t2}\right \|_{2}, \\
\end{split}
\end{equation}
where $\mathbbm{1}_{y^{t1}=y^{t2}}$ is $1$ when $y^{t1}=y^{t2}$ (i.e., sketch and photo are paired), and $0$ elsewhere.
 
Overall, in step 1, we only use $\mathcal{L}_{att}^{s}$ for attribute classification; in step 2, the entire attribute classification loss is
\begin{equation}
\begin{split}
\mathcal{L}_{att}^{s,t1,t2} & = \mathcal{L}_{att}^{s} + \mathcal{L}_{att}^{t1,t2} + \mathcal{L}_{con}^{t1,t2} .
\end{split}
\end{equation}

\subsection{Adversarial Domain Adaptation Learning}

Coupling adversarial training with deep learning introduces a powerful tool to harness domain adaptation. It has been successfully applied to plenty of tasks~\cite{ganin2015unsupervised,wang2018deep,dai2018cross,cao2018partial,zou2019consensus}. 

In IHDA, $\mathcal{X}^{s}$ and $\mathcal{X}^{t1}$ are heterogeneous to $\mathcal{X}^{t2}$. Although $\mathcal{X}^{s}$ and $\mathcal{X}^{t1}$ are closed to homogeneous, there are differences in their illuminations, texture styles and image resolutions. As a result, we accommodate adversarial domain adaptation from the traditional two domains to three domains in this task. 

We build a domain classifier $C_{d}$ to distinguish domains $s$, $t1$ and $t2$. $C_{d}$ consists of three dense layers. For the first two layers, each has $512$ dimensions. The dimension of the output layer is $N_{d}$, which is the number of domains. The network parameters of $C_{d}$ is represented by $\theta_{C_{d}}$. Considering the adversarial domain adaptation is applied, a \textbf{Reverse} Categorical Cross-Entropy Loss is applied to form the loss function:
\begin{equation}
\begin{split}
\hat{\rho}^{s} &=  Softmax \bigg(  C_{d} \big(E_{1}(x^{s}) \big) \bigg), \\
\hat{\rho}^{t1} &=  Softmax \bigg(  C_{d} \big(E_{1}(x^{t1}) \big) \bigg), \\
\hat{\rho}^{t2} &=  Softmax \bigg(  C_{d} \big(E_{2}(x^{t2}) \big) \bigg), \\
\mathcal{L}_{d} &=  \sum_{i=1}^{N_{d}} d^{s}_{i} \log (\hat{\rho}^{s}_{i}) + \sum_{i=1}^{N_{d}} d^{t1}_{i} \log (\hat{\rho}^{t1}_{i}) + \sum_{i=1}^{N_{d}} d^{t2}_{i} \log (\hat{\rho}^{t2}_{i}),\\\\
\end{split}
\end{equation}
where $N_{d}$ stands for the number of domains used for heterogeneous domain adaptation, and $N_{d}=3$ in this work. $d^{s}$, $d^{t1}$ and $d^{t2}$ are one-hot domain labels of $x^s$, $x^{t1}$ and $x^{t2}$, respectively; the subscript $i$ indicates the $i_{th}$ element; $\hat{\rho}^{s}$, $\hat{\rho}^{t1}$ and $\hat{\rho}^{t2}$ are estimated domains of $x^s$, $x^{t1}$ and $x^{t2}$, respectively. 

The adversarial domain adaptation procedure is a two-player game. One player $C_{d}$ intends to maximize $\mathcal{L}_{d}$ so that $d^{s}$, $d^{t1}$ and $d^{t2}$ could be distinguished. Another player is the combination of encoders $E_{1}$ and $E_{2}$, which attempts to minimize $\mathcal{L}_{d}$ and confuse $C_{d}$ by reducing the cross-domain gaps.

The Gradient Reversal Layer ($GRL$)~\cite{ganin2015unsupervised} is applied between the encoder and the domain classifier, which simplifies the two-player adversarial optimization by one step in an iteration.

\subsection{Optimization Goals of IHDA Framework}
By integrating aforementioned Equations, we have the loss functions of step 1 and step 2 as
\begin{equation}
\begin{split}
&\mathcal{L}_{step1} = \mathcal{L}^{s}_{id} + \lambda_1 \mathcal{L}^{s}_{tri} + \lambda_2 \mathcal{L}^{s}_{att},\\
&\mathcal{L}_{step2} = \mathcal{L}^{t1,t2}_{id} + \lambda_1 \mathcal{L}^{t1,t2}_{tri} + \lambda_2 \mathcal{L}^{s,t1,t2}_{att} + \lambda_3 \mathcal{L}_{d},
\end{split}
\label{eq:opt_goal}
\end{equation}
where $\lambda_1$, $\lambda_2$ and $\lambda_3$ are trade-off parameters.

In step 1, the optimization goal is to find the network parameters $\hat{\theta }_{E_{1}}$, $\hat{\theta }_{C^{s}_{id}}$ and $\hat{\theta }_{C_{att}}$ that satisfy

\begin{equation}
\begin{split}
\big( \hat{\theta }_{E_{1}}, \hat{\theta}_{C^{s}_{id}}, \hat{\theta}_{C_{att}} \big) 
&= \underset{\theta_{E_{1}},\theta_{C^{s}_{id}},\theta_{C_{att}}}{\mathrm{argmin}} \mathcal{L}_{step1}.\\
\end{split}
\end{equation}

In step 2, pretrained $\hat{\theta }_{E_{1}}$ and $\hat{\theta}_{C_{att}}$ are resumed. Then $E_{1}$, $E_{2}$, $C^{t1,t2}_{id}$ and $C_{att}$ are simultaneously optimized by minimizing $\mathcal{L}_{step2}$. While $C_{d}$ is optimized by maximizing $\mathcal{L}_{d}$.

The optimization goals are to find the network parameters $\hat{\theta }_{E_{1}}$, $\hat{\theta }_{E_{2}}$, $\hat{\theta }_{C^{t1,t2}_{id}}$, $\hat{\theta }_{C_{att}}$ and $\hat{\theta }_{C_{d}}$ that satisfy
\begin{equation}
\begin{split}
\big( \hat{\theta }_{E_{1}}, \hat{\theta }_{E_{2}}, \hat{\theta}_{C^{t1,t2}_{id}}, \hat{\theta}_{C_{att}} \big) &= \underset{\theta_{E_{1}},\theta_{E_{2}},\theta_{C^{t1,t2}_{id}},\theta_{C_{att}}}{\mathrm{argmin}}\mathcal{L}_{step2}  ,\\
\big(\hat{\theta}_{C_{d}}\big) &=  \underset{\theta_{C_{d}}}{\mathrm{argmax}}
~\mathcal{L}_{d} .\\
\end{split}
\end{equation}

These two optimization goals are jointly updated in each training step.

\subsection{Insight of Optimization Goals}

In IHDA framework, we have $\mathcal{X} \mapsto \mathcal{V}$ (by $E_1$ and $E_2$), $ \mathcal{V} \mapsto \mathcal{Y}$ (by $C^{s}_{id}, C^{t1,t2}_{id}$), and $\mathcal{V} \mapsto \mathcal{Z}$ (by $C_{att}$). We denote $P(v^{s})$ and $P(v^{t})$ as the marginal distributions of embedding features, $P(y^{s}|v^{s})$ and $P(y^{t}|v^{t})$ as the conditional distributions based on identity labels, and $P(z^{s}|v^{s})$ and $P(z^{t}|v^{t})$ as the conditional distributions based on attribute labels, in the source and target domain, respectively.

As theoretically and experimentally analyzed in~\cite{long2013transfer,saito2017asymmetric, long2017deep,zhang2017joint,wang2017balanced,zhao2019learning}, it is insufficient to obtain good embedding representations $\mathcal{V}$ by simply reducing the marginal domain distribution discrepancy. Conditional domain distribution discrepancy should also be considered. 

Therefore, other than coarsely minimizing the marginal domain distribution discrepancy, which is represented by
\begin{equation}
\begin{split}
\textbf{min}\{\left \|P(v^{s})-P(v^{t})\right \|\},
\label{eq:ideal_goal}
\end{split}
\end{equation}
we aim to jointly reduce marginal domain discrepancy and conditional domain distribution discrepancy: 
\begin{equation}
\begin{split}
\left\{\begin{matrix}
&\textbf{min}\{\left \|P(v^{s})-P(v^{t})\right \|\},\\
&\textbf{min}\{\left \|P(y^{s}|v^{s})-P(y^{t}|v^{t})\right \|\}.
\end{matrix}\right.
\label{eq:ideal_goal_2}
\end{split}
\end{equation}

Ideally, we should minimize $\left \|P(y^{s}|v^{s})-P(y^{t}|v^{t})\right \|$, however, as we have analysed, in instance-level retrieval tasks, we have $\mathcal{Y}^{s} \cap \mathcal{Y}^{t} = \varnothing$, it is challenging to minimize $\left \|P(y^{s}|v^{s})-P(y^{t}|v^{t})\right \|$. 

Since attribute learning has a main training objective on attributes but also learn representations that can identify individuals as a side effect, we substitute $\left \|P(y^{s}|v^{s})-P(y^{t}|v^{t})\right \|$ with $\left \|P(z^{s}|v^{s})-P(z^{t}|v^{t})\right \|$, where $z$ is the attribute and we have $\mathcal{Z}^{s}=\mathcal{Z}^{t}$. Then we obtain the following optimization goal:
\begin{equation}
\begin{split}
\left\{\begin{matrix}
&\textbf{min}\{\left \|P(v^{s})-P(v^{t})\right \|\} \Leftrightarrow 
\mathcal{L}_{d}, \\
&\textbf{min}\{\left \|P(z^{s}|v^{s})-P(z^{t}|v^{t})\right \|\}
\Leftrightarrow \mathcal{L}^{s,t1,t2}_{att}.
\end{matrix}\right.
\label{eq:real_goal}
\end{split}
\end{equation}

They are identical to the last two items of $\mathcal{L}_{step2}$ (see Equation~\ref{eq:opt_goal}).
We minimize the conditional distribution discrepancy $\mathcal{L}^{s,t1,t2}_{att}$ by a semi-supervised learning approach~\cite{grandvalet2005semi,saito2019semi}. Be the same as balance factors in Balanced Distribution Adaptation~\cite{wang2017balanced}, $\lambda_2$ and $\lambda_3$ are used to adjust the importance reducing the marginal distribution discrepancy and conditional distribution discrepancy.

Consequently, we approach jointly optimize the marginal distribution and conditional distribution in our IHDA. 
We also experimentally show the effectiveness of jointly optimize the marginal distribution and conditional distribution in DA in ablation studies.

\section{Experiments}
\label{sec:experiments}

\subsection{Datasets}

In our experiments, source datasets include \textbf{UT-Zap50K}~\cite{semjitter}, \textbf{CelebFaces}~\cite{liu2015faceattributes}, and \textbf{Market-1501}~\cite{zheng2015scalable}, their corresponding target datasets are \textbf{QMUL-Shoes}~\cite{yu2016sketch}, \textbf{IIIT-D Viewed Sketch}~\cite{bhatt2012memetically} and \textbf{PKU-Sketch}~\cite{pang2018cross}, respectively. Since sketches do not contain color information, we select non-color attributes from the annotated source to form the shared label space $\mathcal{Z}^{s}=\mathcal{Z}^{t1,t2}$.

1) The UT-Zap50K dataset is a large-scale shoe photo dataset consisting of 50K images and around 20K identities. We select all non-color attributes to form the shared label space, such as ``Wide Toe'', ``Snip Toe'', ``Toggle Closure'', and so on. The QMUL-Shoes dataset contains 419 sketch-to-photo pairs of shoes, in which 304 pairs are for training and 115 pairs are for testing. Since sketches are collected from non-professional drawers, it encounters more challenges to recognize sketch-to-photo pairs than others. Although the target QMUL-Shoes also includes shoes attributes, they are different from the ones defined in UT-Zap50K dataset. To perform supervised learning on target attributes (e.g.,~\cite{liu2018deep}), we need to add new attributes to the target, which is laborious. Our method takes advantage of unsupervised DA and does not use labeled attributes in the target dataset.

2) The CelebFaces dataset is a large-scale face photo dataset with around 10K identities, 202K face images, and 40 binary attributes. We select all non-color attributes to form the shared label space, such as ``Wearing-Hat'', ``Bald'', ``Eyeglasses'', and so on. The IIIT-D Viewed Sketch dataset includes 238 sketch-to-photo pairs of faces, in which the sketches are drawn by professional artists. Unlike our source dataset CelebFaces, attribute annotation is not available in IIIT-D Viewed Sketch dataset. We take the same evaluation protocols as~\cite{wu2018light,deng2019residual} proposed.

3) The Market-1501 dataset is a large-scale person photo dataset with around 1.5K identities, 32K person images. In work~\cite{lin2019improving,wang2019learning}, 27 binary attributes are annotated for Market-1501 dataset. We select all non-color attributes to form the shared label space, as ``gender'', ``hair'', ``length of sleeve'', ``length of lower-body clothing'', ``style of clothing'', ``hat'', ``age'', and ``bag'' (``hand bag'' and ``pack bad'' are regarded as ``bag''). The PKU-Sketch dataset consists of samples of 200 persons, in which each person has one sketch and two photos taken from disjoint cameras. Besides, there is no attribute annotation for it. We take the evaluation protocols~\cite{pang2018cross}, and 150 identities are randomly chosen for training while the rest 50 are used for testing. We report average values of 10 times experiments as the final results.

\emph{Note that, we utilize the whole source dataset and a small portion of the annotated target training set for the training, while the target testing set is untouched in the training process}.

\subsection{Experimental Settings}
Following the procedure in Fig.~\ref{fig:model}, we start with pre-training the $E_{1}$ branch on the source data with $60$ epochs in step 1. The batch size is $64$. Then, in step 2, by reloading the pre-trained $\theta_{E_{1}}$ and $\theta_{C_{att}}$ weights, we co-train the whole IHDA network on both source and target training data with another $60$ epochs. The batch size is $96$ (32 sketch-to-photo pairs from the target dataset and 32 samples from the source dataset). We choose Adam optimizer~\cite{kingma2014adam} for the training and the initial learning rate is set to be $1\times10^{-4}$. To properly utilize pre-trained network parameters, we apply a warming-up learning rate schedule, which is modified from~\cite{Luo_2019_Strong_TMM}, to adjust the learning rate. We set $\lambda_1=1$, $\lambda_2=0.1$ and $\lambda_3=0.1$ to balance the whole network. Note that the input image size for shoe, face, and person samples are $96\times96$, $144\times128$, and $384\times128$, respectively. In the inference stage, a $L_2$ Normalization is applied to each embedding feature before calculating their relative distances.

\subsection{Comparison with Methods of Sketch-to-photo Retrieval}

We compare our method with state-of-the-art methods of sketch-to-photo retrieval on three types of target datasets, respectively for the sketch-to-photo retrieval of shoes, faces, and persons. 

For the QMUL-Shoes dataset, we compare our method with Triplet SN~\cite{yu2016sketch}, MAR-FBIR~\cite{song2016deep}, CD-AFL~\cite{pang2018cross}, and DSSA Triplet~\cite{song2017deep}. They are pre-trained on external datasets TU Berlin Sketch~\cite{eitz2012sbsr} and Edge-style ImageNet~\cite{russakovsky2015imagenet}. Our model takes the UT-Zap50K dataset as the source dataset since shoe attributes are annotated in it. The rank-1 and rank-10 re-identification accuracy of each method are reported in TABLE~\ref{t:results_QMUL}.

\tabcolsep=4pt
\begin{table}[!ht]
\centering
\caption{The performance of sketch-to-photo retrieval on the QMUL-Shoes dataset.}
\resizebox{\linewidth}{!}{
\smallskip
\begin{tabular}{l|c|cc}
\toprule
\textbf{Model}      &\textbf{w/\ external Data}    & \textbf{R1} & \textbf{R10}    \\ 
\midrule
\begin{tabular}[c]{@{}c@{}} Triplet SN~\cite{yu2016sketch} \\(pre-train w/o att + fine-tune) \end{tabular}
& \begin{tabular}[c]{@{}c@{}} TU Berlin Sketch  \\\&Edge-style ImageNet \end{tabular}
&39.1  &87.8 \\ 
\begin{tabular}[c]{@{}c@{}} MAR-FBIR~\cite{song2016deep} \\(pre-train w/o att + fine-tune) \end{tabular}
& \begin{tabular}[c]{@{}c@{}} TU Berlin Sketch  \\\&Edge-style ImageNet \end{tabular}
&50.4  &91.3 \\
\begin{tabular}[c]{@{}c@{}} CD-AFL~\cite{pang2018cross}  \\(pre-train w/o att + fine-tune) \end{tabular}
& \begin{tabular}[c]{@{}c@{}} TU Berlin Sketch  \\\&Edge-style ImageNet \end{tabular}
&56.4  &92.6 \\ 
\begin{tabular}[c]{@{}c@{}} DSSA Triplet~\cite{song2017deep}  \\(pre-train w/o att + fine-tune) \end{tabular}
& \begin{tabular}[c]{@{}c@{}} TU Berlin Sketch  \\\&Edge-style ImageNet \end{tabular}
&61.7  &94.8 \\  \midrule 
\begin{tabular}[c]{@{}c@{}} Our IHDA  \\(Instance-level heterogeneous DA w/ att) \end{tabular}   &UT-Zap50K & \textbf{68.7} &\textbf{95.7}\\
\bottomrule
\end{tabular}
}
\label{t:results_QMUL}
\end{table}

For the IIIT-D Viewed Sketch dataset, we compare our method with Deep Face~\cite{parkhi2015deep}, Light CNN~\cite{wu2018light}, CDL~\cite{wu2018coupled}, and RCN~\cite{deng2019residual}. The CelebFaces dataset is used in their pre-training. Our model takes the CelebFaces dataset as the source dataset and uses the IHDA framework. The rank-1 re-identification accuracy of each method is listed in TABLE~\ref{t:results_IIITD}.
\tabcolsep=4pt
\begin{table}[!ht]
\centering
\caption{The performance of sketch-to-photo retrieval on the IIIT-D Viewed Sketch dataset.}
\resizebox{\linewidth}{!}{
\begin{tabular}{l|c|c}
\toprule
\textbf{Model}      & \textbf{w/\ external Data}  & \textbf{R1}   \\ 
\midrule
\begin{tabular}[c]{@{}c@{}} Deep Face~\cite{parkhi2015deep}  \\(pre-train w/o att + fine-tune) \end{tabular}
& CelebFaces & 80.9  \\  
\begin{tabular}[c]{@{}c@{}} Light CNN~\cite{wu2018light}  \\(pre-train w/o att + fine-tune) \end{tabular}
& CelebFaces & 84.0  \\
\begin{tabular}[c]{@{}c@{}} CDL~\cite{wu2018coupled} \\(pre-train w/o att + fine-tune) \end{tabular}
& CelebFaces & 85.4\\ 
\begin{tabular}[c]{@{}c@{}} RCN~\cite{deng2019residual} \\(pre-train w/o att + fine-tune) \end{tabular}
& CelebFaces & 90.3\\
\midrule
\begin{tabular}[c]{@{}c@{}} Our IHDA  \\(Instance-level heterogeneous DA w/ att) \end{tabular}   & CelebFaces & \textbf{95.7}\\  
\bottomrule
\end{tabular}
}
\label{t:results_IIITD}
\end{table}

For the PKU-Sketch dataset, we compare our method with Triplet SN~\cite{yu2016sketch}, GN Siamese~\cite{SangkloyBHH16}, and CD-AFL~\cite{pang2018cross}. Among them, GN Siamese and CD-AFL are pre-trained on the Market-1501 dataset. Note that CD-AFL only applied adversarial supervised DA on PKU-Sketch dataset. When considering Market-1501 dataset together, it simply applied pre-training and fine-tuning. Our model takes the Market-1501 dataset as the source dataset and uses the IHDA framework. The rank-1, rank-5, rank-10 and rank-20 re-identification accuracy of each method are reported in TABLE~\ref{t:results_PKU-Sketch}.

\tabcolsep=4pt
\begin{table}[!ht]
\centering
\caption{The performance of sketch-to-photo retrieval on the PKU-Sketch dataset.}
\resizebox{1\linewidth}{!}{
\smallskip
\begin{tabular}{l|c|cccc}
\toprule
\textbf{Model}      & \textbf{w/\ external Data}  & \textbf{R1} & \textbf{R5} & \textbf{R10} & \textbf{R20}    \\ \midrule
\begin{tabular}[c]{@{}c@{}} Triplet SN~\cite{yu2016sketch} \\(pre-train w/o att + fine-tune) \end{tabular} & $\times$ & 9.0   & 26.8  & 42.2   & 65.2  \\
\begin{tabular}[c]{@{}c@{}} GN Siamese~\cite{SangkloyBHH16} \\(Ppe-train w/o att + fine-tune) \end{tabular} & Market-1501& 28.9  & 54.0  & 62.4   & 78.2  \\  
\begin{tabular}[c]{@{}c@{}} CD-AFL~\cite{pang2018cross} \\(pre-train w/o att + fine-tune) \end{tabular} & Market-1501 &  34.0  & 56.3  & 72.5 & 84.7 \\   \midrule
\begin{tabular}[c]{@{}c@{}} Our IHDA  \\(Instance-level heterogeneous DA w/ att) \end{tabular}   & Market-1501 & \textbf{85.6} &\textbf{94.8} &\textbf{98.0} &\textbf{100.0}\\ 
\bottomrule
\end{tabular}
}
\label{t:results_PKU-Sketch}
\end{table}

\begin{figure}[!htb]
\centering
  \includegraphics[width=1\columnwidth]{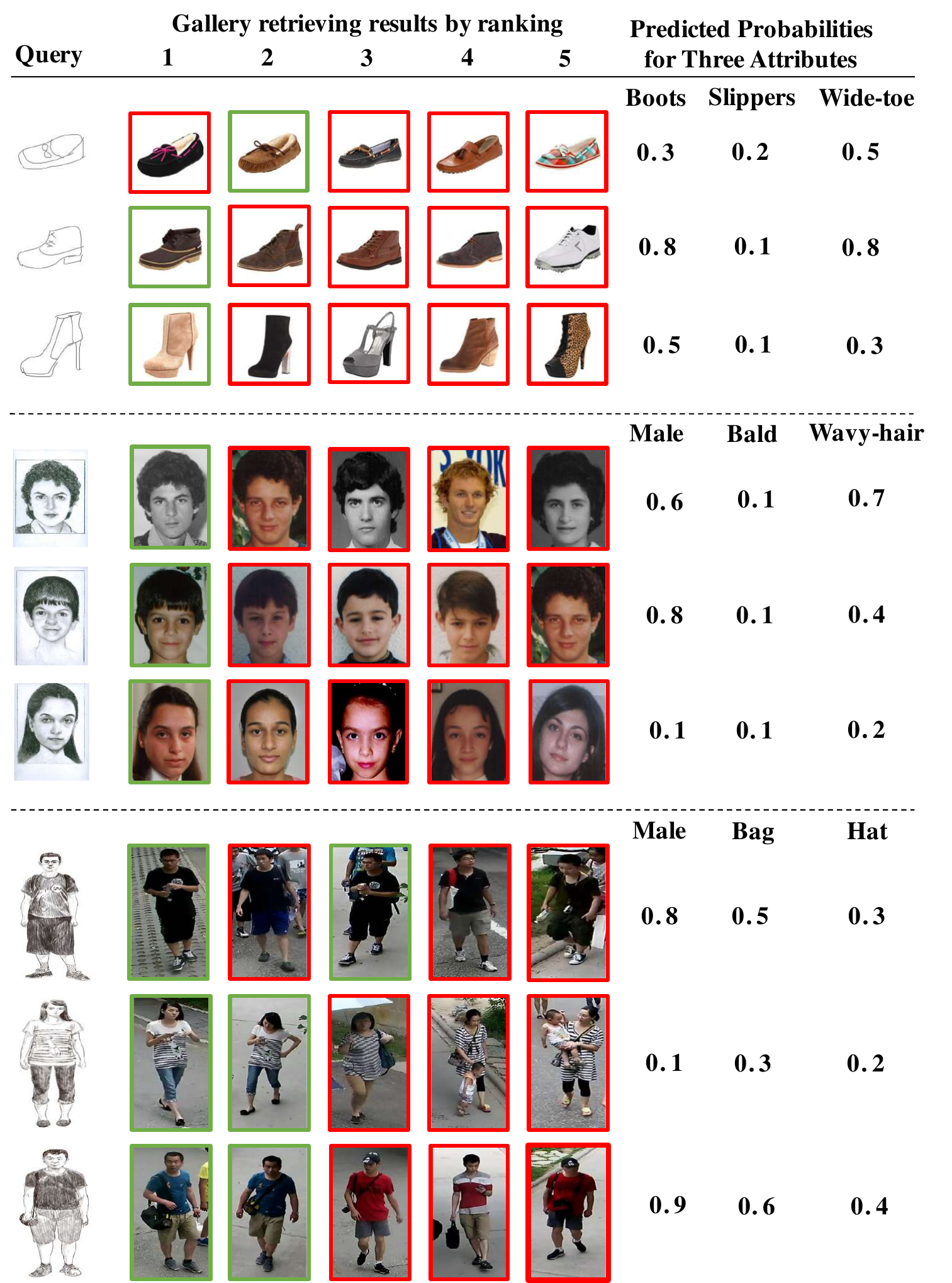}
  \caption{Visualization of retrieving results on QMUL-Shoes, IIIT-D, and PKU-Sketch datasets. In the same row, gallery images with green bounding boxes are identical to the query image, while others with the red bounding box are different identities. Predicted probabilities are shown for three attributes here.}
  \label{fig:visual}
\end{figure}

On QMUL-Shoes dataset, using IHDA framework outperforms the previous state-of-the-art method by $7.0\%$ and reaches $68.7\%$ in rank-1 retrieving accuracy. A $5.4\%$ gain of rank-1 retrieving accuracy is made on IIIT-D Viewed Sketch dataset, with a value as high as $95.7\%$. On PKU-Sketch dataset, IHDA surpasses the previous state-of-the-art method up to $51.6\%$ and obtains $85.6\%$ in rank-1 retrieving accuracy. This is not happened by coincidence and we will go deep into the essential mechanism of IHDA in ablations studies.

The improvement in PKU-Sketch dataset is more significant than the other two datasets. We consider there are two main reasons: (1) Sketches of QMUL-Shoes dataset are drawn by amateurs while sketches of IIIT-D and PKU-Sketch datasets are drawn by artists with high qualities and more details (see Fig.~\ref{fig:visual}), we can achieve high performance on IIIT-D and PKU-Sketch datasets. (2) The body poses within a body sketch-to-photo pair could be remarkably different, which causes larger distortion than face sketch-to-photo pairs. Therefore, the fine-tuning strategy can reach a good performance on the IIIT-D dataset, but on the more challenging PKU-Sketch dataset, our IHDA shows its superiority.

In Fig.~\ref{fig:visual}, we illustrate several retrieving results on target testing data. It can be observed that each retrieved photo has a similar appearance and structure as the query sketch. This proves that the joint heterogeneous embedding is well learned in our IHDA framework. For some abstract sketches, it is even difficult to identify which is the correct corresponding photo by eyes. This explains why our IHDA framework is hindered to further improve sketch-to-photo retrieving performance.

\subsection{Comparison of Using Different Source Datasets and Training Strategies }

Due to the lack of annotations, most of sketch-photo retrieval works~\cite{song2016deep, pang2018cross, song2017deep, parkhi2015deep, wu2018light, wu2018coupled, deng2019residual,SangkloyBHH16} utilize external datasets to pre-train and then fine-tune their model on the target dataset. However, there two questions worth to be further explored: 1) though unrelated sketch-photo source data (e.g., Sketchy~\cite{sangkloy2016sketchy}) cannot fit our complete IHDA framework, could an incomplete IHDA framework obtain satisfactory results by using it as the source dataset? 2) since attribute learning is applied in our IHDA framework, could it also improve the pre-training and fine-tuning framework? To answer these questions, we construct three baselines by using part of our IHDA framework components and perform the same training strategy on three experimental datasets.

Specifically, in Baseline A, we conduct experiments by using Sketchy dataset~\cite{sangkloy2016sketchy} as the source dataset, and QMUL-Shoes, IIIT-D Viewed Sketch, and PKU-Sketch as the target datasets. Referring to the training process in Fig.~\ref{fig:model}, we first pre-train two encoders (for photo and sketch separately) using Sketchy in Step 1, and then fine-tune both encoders using target datasets in Step 2. In Baseline B and C, whose components and source dataset are the same as IHDA framework, excepting that attribute learning is removed or only used in Step 1. The results are shown in TABLE \ref{t:results_QMUL_ablation}, TABLE \ref{t:results_IIITD_ablation}, and TABLE \ref{t:results_PKU-Sketch_ablation}.

\tabcolsep=4pt
\begin{table}[!h]
\centering
\caption{The source dataset setting and sketch-to-photo retrieval performance on the QMUL-Shoes dataset.}
\resizebox{\linewidth}{!}{
\smallskip
\begin{tabular}{l|c|cc}
\toprule
\textbf{Model}      &\textbf{w/\ external Data}    & \textbf{R1} & \textbf{R10}    \\ 
\midrule
\begin{tabular}[c]{@{}c@{}} Our Baseline A  \\(pre-train w/o att + fine-tune) \end{tabular} & Sketchy &59.2 &93.3 \\ 
\begin{tabular}[c]{@{}c@{}} Our Baseline B  \\(pre-train w/o att + fine-tune) \end{tabular}   & UT-Zap50K &57.5 &92.8\\ 
\begin{tabular}[c]{@{}c@{}} Our Baseline C  \\(pre-train w/ att + fine-tune) \end{tabular}  &UT-Zap50K   & 59.0 &93.1\\  \midrule
\begin{tabular}[c]{@{}c@{}} Our IHDA  \\(Instance-level heterogeneous DA w/ att) \end{tabular}   &UT-Zap50K & \textbf{68.7} &\textbf{95.7}\\
\bottomrule
\end{tabular}
}
\label{t:results_QMUL_ablation}
\end{table}

\tabcolsep=4pt
\begin{table}[!h]
\centering
\caption{The source dataset setting and sketch-to-photo retrieval performance on the IIIT-D Viewed Sketch dataset.}
\resizebox{\linewidth}{!}{
\begin{tabular}{l|c|c}
\toprule
\textbf{Model}      & \textbf{w/\ external Data}  & \textbf{R1}   \\ 
\midrule
\begin{tabular}[c]{@{}c@{}} Our Baseline A  \\(pre-train w/o att + fine-tune) \end{tabular} & Sketchy &90.5 \\ 
\begin{tabular}[c]{@{}c@{}} Our Baseline B  \\(pre-train w/o att + fine-tune) \end{tabular}   & CelebFaces &88.3 \\ 
\begin{tabular}[c]{@{}c@{}} Our Baseline C  \\(pre-train w/ att + fine-tune) \end{tabular}  &CelebFaces  & 88.6 \\  \midrule
\begin{tabular}[c]{@{}c@{}} Our IHDA  \\(Instance-level heterogeneous DA w/ att) \end{tabular}   & CelebFaces & \textbf{95.7}\\  
\bottomrule
\end{tabular}
}
\label{t:results_IIITD_ablation}
\end{table}

\tabcolsep=4pt
\begin{table}[!ht]
\centering
\caption{The source dataset setting and sketch-to-photo retrieval performance on the PKU-Sketch dataset.}
\resizebox{\linewidth}{!}{
\smallskip
\begin{tabular}{l|c|cccc}
\toprule
\textbf{Model}      & \textbf{w/\ external Data}  & \textbf{R1} & \textbf{R5} & \textbf{R10} & \textbf{R20}    \\
\midrule
\begin{tabular}[c]{@{}c@{}} Our Baseline A  \\(pre-train w/o att + fine-tune) \end{tabular} & Sketchy & 61.0      &82.6 &93.6       &95.2 \\ 
\begin{tabular}[c]{@{}c@{}} Our Baseline B  \\(pre-train w/o att + fine-tune) \end{tabular}   & Market-1501 &58.8       &78.0       &90.0       &94.2 \\ 
\begin{tabular}[c]{@{}c@{}} Our Baseline C  \\(pre-train w/ att + fine-tune) \end{tabular}  &Market-1501  &59.4       &79.2       &91.2       &95.8 \\  \midrule
\begin{tabular}[c]{@{}c@{}} Our IHDA  \\(Instance-level heterogeneous DA w/ att) \end{tabular}   & Market-1501 & \textbf{85.6} &\textbf{94.8} &\textbf{98.0} &\textbf{100.0}\\ 
\bottomrule
\end{tabular}
}
\label{t:results_PKU-Sketch_ablation}
\end{table}

The results of Baseline A consistently demonstrate when the naive pre-training and fine-tuning strategy is applied, using Sketchy dataset for pre-training leads to better performance than using a homogeneous dataset. Since Sketchy dataset contains sketch and photo pairs, using it in pre-training might improve low-level feature representation in both encoders. However, samples in Sketchy dataset and our target datasets may not belong to the same category, not matter the same identity. Such a content divergence hinders the model's capability on transferring the high-level semantic knowledge to target datasets. To some extent, an ideal source dataset might satisfy: 1) including sketch-photo pairs (Sketchy dataset); 2) including different identities but the same category as the target dataset (our selected source datasets). However, it is difficult to find such an ideal source dataset for all target datasets, we have to make trade-offs. Luckily, our proposed IHDA framework significantly improves the performance of target datasets by assuming the availability of attributes but dropping the need for category-matched sketch-photo data.

The performances of Baseline B and C are similar, which indicates that \emph{solely including attribute learning in existing approaches \cite{pang2018cross,song2017deep,deng2019residual} may not remarkably improve the retrieval performance on the target data}. More than using attributes in pre-training, our IHDA uses attributes to form a shared label space and perform semi-supervised attribute learning to intermediately reduce the domain gaps in terms of marginal distributions and conditional distributions. As a result, our IHDA can significantly improve the performance in three target datasets.

\subsection{Comparison with Methods of Photo-to-sketch Retrieval}

Although the sketch-to-photo retrieval (query: sketch; gallery: photo) is a dominant application scenario, we perform an inverse experiment to investigate the performance of photo-to-sketch retrieval (query: photo; gallery: sketch) on three benchmarks used in our paper. The results are shown in Tables \ref{t:results_QMUL_Shoes_p2s}, \ref{t:results_IIIT-D_p2s}, and \ref{t:results_PKU-Sketch_p2s} (Note, some unpublished codes are re-implemented by us, bias could exist). 

\tabcolsep=4pt
\begin{table}[!ht]
\centering
\resizebox{\linewidth}{!}{
\smallskip
\begin{tabular}{l|c|ccc}
\toprule
\textbf{Model}      &\textbf{w/\ external Data}    & \textbf{R1} & \textbf{R10} & \textbf{R20}    \\ 
\midrule
\begin{tabular}[c]{@{}c@{}} DSSA Triplet~\cite{song2017deep}  \\(pre-train w/o att + fine-tune) \end{tabular}
& \begin{tabular}[c]{@{}c@{}} TU Berlin Sketch  \\\&Edge-style ImageNet \end{tabular}&61.9 &95.1 &98.6\\\midrule
\begin{tabular}[c]{@{}c@{}} Our IHDA  \\(Instance-level heterogeneous DA w/ att) \end{tabular}   & UT-Zap50K &\textbf{69.6} &\textbf{97.4} &\textbf{99.1}\\
\bottomrule
\end{tabular}
}
\caption{The performance of photo-to-sketch retrieval on the QMUL-Shoes dataset.}
\label{t:results_QMUL_Shoes_p2s}
\end{table}

\tabcolsep=4pt
\begin{table}[!ht]
\centering
\resizebox{\linewidth}{!}{
\begin{tabular}{l|c|ccc}
\toprule
\textbf{Model}      &\textbf{w/\ external Data}    & \textbf{R1} & \textbf{R10} & \textbf{R20}    \\ 
\midrule
\begin{tabular}[c]{@{}c@{}} RCN~\cite{deng2019residual} \\(pre-train w/o att + fine-tune) \end{tabular}
& CelebFaces & 90.8 & 95.2 & 97.8 \\ \midrule
\begin{tabular}[c]{@{}c@{}} Our IHDA  \\(Instance-level heterogeneous DA w/ att) \end{tabular} & CelebFaces & \textbf{96.2}&\textbf{98.6} &\textbf{99.2}\\ 
\bottomrule
\end{tabular}
}
\caption{The performance of photo-to-sketch retrieval on the IIIT-D Viewed Sketch dataset.}
\label{t:results_IIIT-D_p2s}
\end{table}

\tabcolsep=4pt
\begin{table}[!ht]
\centering
\resizebox{\linewidth}{!}{
\smallskip
\begin{tabular}{l|c|ccc}
\toprule
\textbf{Model}   &\textbf{w/\ external Data} & \textbf{R1}  & \textbf{R10} & \textbf{R20}    \\
\midrule
\begin{tabular}[c]{@{}c@{}} CD-AFL~\cite{pang2018cross} \\(pre-train w/o att + fine-tune) \end{tabular} & Market-1501  & 37.6  & 76.2 & 92.8 \\  \midrule 
\begin{tabular}[c]{@{}c@{}} Our IHDA  \\(Instance-level heterogeneous DA w/ att) \end{tabular}   &  Market-1501  & \textbf{88.2}  & \textbf{100.0}   & \textbf{100.0}  \\  
\bottomrule
\end{tabular}
}
\caption{The performance of photo-to-sketch retrieval on the PKU-Sketch dataset.}
\label{t:results_PKU-Sketch_p2s}
\end{table}

The photo-to-sketch retrieval performance is similar to sketch-to-photo retrieval performance, which indicates that distinguishable embedding is obtained for sketches and photos in a symmetrical way.

\subsection{Comparison with Methods of Instance-level Domain Adaptation on Sketch-to-photo Retrieval Tasks}

Existing instance-level domain adaption methods mainly perform their experiments on cross-dataset photo (i.e., homogeneous) retrieval tasks. In this part, we compare our IHDA framework with two open-source works~\cite{ge2020mutual,song2020unsupervised} on PKU-Sketch dataset. The results are shown in Tables \ref{t:results_PKU-Sketch_da}.

\tabcolsep=4pt
\begin{table}[!ht]
\centering
\caption{The performance of sketch-to-photo retrieval on the PKU-Sketch dataset (compared with DA methods).}
\resizebox{1\linewidth}{!}{
\smallskip
\begin{tabular}{l|c|cccc}
\toprule
\textbf{Model}      & \textbf{w/\ external Data}  & \textbf{R1} & \textbf{R5} & \textbf{R10} & \textbf{R20}    \\ \midrule
\begin{tabular}[c]{@{}c@{}} MMT~\cite{ge2020mutual} \\(Instance-level homogeneous DA w/o att) \end{tabular} & Market-1501 &  42.8  & 62.4  & 71.2 & 88.2 \\   \midrule 
\begin{tabular}[c]{@{}c@{}} UDA-Reid~\cite{song2020unsupervised} \\(Instance-level homogeneous DA w/o att) \end{tabular} & Market-1501 &  40.4  & 60.0  & 72.8 & 90.4 \\   \midrule
\begin{tabular}[c]{@{}c@{}} Our incomplete IHDA  \\(Instance-level heterogeneous DA w/o att) \end{tabular}
& Market-1501 & 70.4 & 82.0 & 92.6 & 96.2 \\ \midrule
\begin{tabular}[c]{@{}c@{}} Our complete IHDA  \\(Instance-level heterogeneous DA w/ att) \end{tabular}
& Market-1501 & \textbf{85.6} &\textbf{94.8} &\textbf{98.0} &\textbf{100.0}\\ 
\bottomrule
\end{tabular}
}
\label{t:results_PKU-Sketch_da}
\end{table}

Though MMT~\cite{ge2020mutual} and UDA-Reid~\cite{song2020unsupervised} have achieved great successes on cross-dataset homogeneous retrieval tasks (\textit{e.g.}, synthetic/real photo), heterogeneous (\textit{e.g.}, sketch-to-photo) domain gaps may be out of the capabilities of their models. Even without attribute learning, our incomplete IHDA still can significantly outperform MMT and UDA-Reid on the PKU-Sketch dataset. In particular, their domain adaptation is specifically designed to improve feature representation in cross-dataset photo (homogeneous) retrieval tasks, and thus a single encoder is applied. However, we consider that applying domain adaption on sketch-to-photo (heterogeneous) retrieval task is different. To customize instance-level domain adaptation to the heterogeneous case, our IHDA framework applies two encoders for sketch and photo separately. Besides, our domain adaptation consists of three domains, as $\mathcal{D}^{s}$, $\mathcal{D}^{t1}$ and $\mathcal{D}^{t2}$, and corresponding objective functions are designed for jointly reducing domain gaps within them. Moreover, unlike MMT and UDA-Reid that need to include the target testing set in their training, our IHDA framework keeps the target testing set untouched during the training process, which demonstrates its better generalization.

\subsection{Attribute Selection}

Since the source datasets are not originally designed for the target datasets, some attributes of source datasets may not suitable for the target dataset. We would like to highlight that task-irrelevant attributes, which is against our assumption $\mathcal{Z}^{s}=\mathcal{Z}^{t1,t2}$, should be excluded. in terms of sketch-to-photo retrieval task, color-related attributes are task-irrelevant. As all attributes are defined at the semantic level, excluding color attributes of the source dataset is \textbf{deterministic} rather than heuristic.

What will happen if we select color attributes to apply IHDA framework on the sketch-to-photo retrieval task? Taking Market-1501 and PKU-Sketch datasets as an example, we do experiments to prove that using color attributes can cause a negative transfer. As TABLE~\ref{t:results_PKU-Sketch_color} shows, the sketch-photo retrieval performance significantly decreases after adding color attributes. The color attributes may work as a noise, make the model has no incentive to learn the real relevant semantic information between source and target domains. Compared with manually annotating sketch-photo datasets, it worth spending a few seconds to exclude color attributes at the beginning.

   \tabcolsep=8pt
   \begin{table}[!ht]
   \centering
   \resizebox{\linewidth}{!}{
   \smallskip
     \begin{tabular}{l|cccc}
   \toprule
   \textbf{Model}  & \textbf{R1} & \textbf{R5} & \textbf{R10} & \textbf{R20}    \\
   \midrule
   IHDA (w/o color attributes)  & 85.6  & 94.8  & 98.0   & 100.0  \\  
   IHDA (w/ color attributes)  & 54.4  & 76.2  & 88.6   & 90.2  \\
   \bottomrule
   \end{tabular}
   }
   \caption{Comparison of using color attributes on PKU-Sketch dataset.}
   \label{t:results_PKU-Sketch_color}
   \end{table}

\begin{figure}[!ht]
    \centering
    \includegraphics[width=\textwidth]{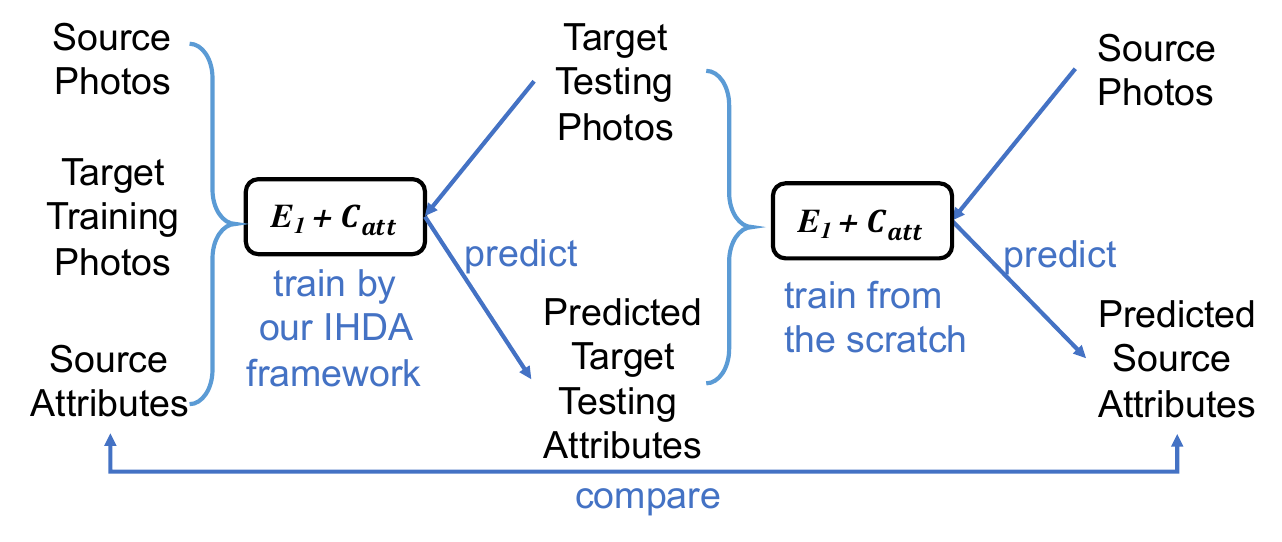}
    \caption{The procedure of selecting attributes for IHDA framework.}
  \label{fig:select_att}
\end{figure}

Aside from color attributes, there are several non-color attributes could be selected. Since we may not have the ground-truth attributes on the target dataset, it is challenging to apply traditional feature selection methods~\cite{guyon2003introduction} to select the most useful ones in our task. As an alternative approach, we assume that if the semantic information of selected attributes can be well learned in the target domain, a new attribute classifier, which is trained from scratch using predicted target attributes, should correctly recognize the attribute in the source domain. Based on this assumption, we propose a novel procedure to select attributes as Fig.~\ref{fig:select_att} shows. First, we utilize all available non-color attributes to run our IHDA framework with the default training setting. Then, we estimate attributes for target testing photos, which are untouched during the training process. After that, target testing photos and corresponding predicted attributes are used to train the photo encoder (i.e., $E_{1}$) and attribute classifier (i.e., $C_{att}$) from the scratch. Finally, we estimate the source attributes and compare them with the ground truth. The comparison results are used for attribute selection.

We use the PKU-Sketch dataset to demonstrate such an attribute selection procedure. The attribute recognition accuracy on the source dataset is reported in TABLE~\ref{t:PKU-Sketch_att_selection}.

\tabcolsep=4pt
\begin{table}[!ht]
\centering
\caption{ Attribute recognition accuracy on source dataset (trained on target photo training set). ``L.slv'', ``L.low'', ``S.clth'' denote ``length of sleeve'', ``length of lower-body clothing'', ``style of clothing'', resp.}
\resizebox{\linewidth}{!}{
\smallskip
\begin{tabular}{l|cccccccc}
\toprule
\textbf{Attributes}&\textbf{L.slv}  &\textbf{L.low} & \textbf{bag} &\textbf{gender}  &\textbf{hair}  &\textbf{S.clth}  &\textbf{hat} &\textbf{age}  \\ \midrule
\textbf{Accuracy (\%)} &81.6 & 75.9 &73.7 & 72.1 &70.5  &68.4 &62.1 & 37.5  \\
\bottomrule
\end{tabular}
}
\label{t:PKU-Sketch_att_selection}
\end{table}

Based on our assumption, the attribute with a higher recognition accuracy might contribute more in our IHDA framework. To verify this assumption, we further select two groups of attributes, as ``L.slv, L.low, bag'' and ``S.clth, hat, age'', to run IHDA framework. The results in TABLE~\ref{t:results_PKU-Sketch_diff_att} agree with our assumption.

  \tabcolsep=8pt
  \begin{table}[!ht]
  \centering
  \resizebox{\linewidth}{!}{
  \smallskip
  \begin{tabular}{l|cccc}
  \toprule
  \textbf{Model}  & \textbf{R1} & \textbf{R5} & \textbf{R10} & \textbf{R20}    \\
  \midrule
  IHDA (w/ L.slv, L.low, bag)  & 84.2  & 94.4  & 97.8   & 100.0  \\  
  IHDA (w/ S.clth, hat, age)  & 80.0  & 93.2  & 96.6   & 98.0  \\
  \bottomrule
  \end{tabular}
  }
  \caption{The sketch-to-photo retrieval performance by using different attributes on PKU-Sketch dataset.}
  \label{t:results_PKU-Sketch_diff_att}
  \end{table}

Although we can select important attributes through the above method, it is inefficient. Considering that we have chosen closely related source datasets for target datasets, non-color attributes may generally be shared between them and the total number of attributes should not be a large number. Could we just use all non-color attributes? 

\begin{figure}[h!]
\centering 
\includegraphics[width=0.65\columnwidth]{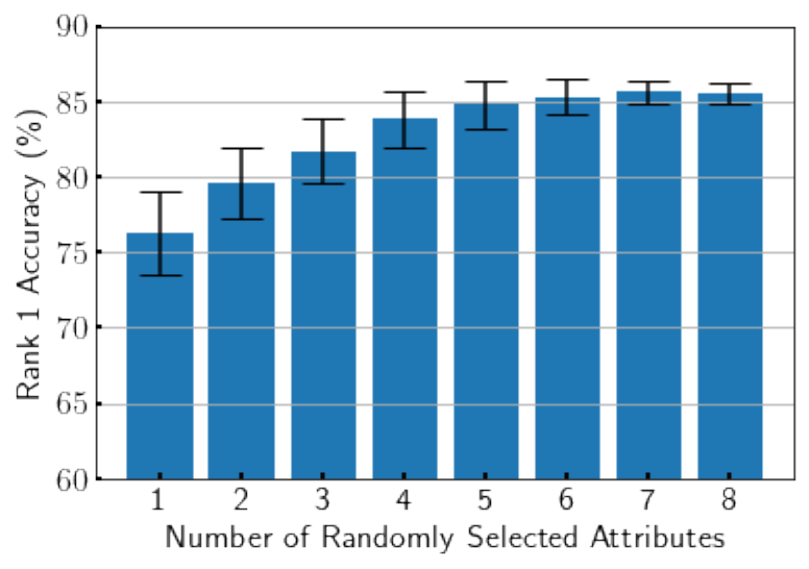}
\caption{Analysis on attribute selection.}
\label{fig:ablation_att}
\end{figure}

We do another experiment on PKU-Sketch dataset to explore this question. Within all non-color attributes ($8$ in total), we randomly select $k$ (range from $1$ to $8$) of them and repeat for $10$ times to obtain average rank-1 value and uncertainty. As Fig. \ref{fig:ablation_att} shows, 
in the beginning, with more attributes, the rank-1 value is steadily increasing while the uncertainty is decreasing. When more than $6$ attributes are used, there is a small difference in the rank-1 value and uncertainty. It suggests that, within a certain range, increasing the number of attributes helps to improve the performance of IHDA framework. However, above a certain number, adding more shared attributes may not significantly affect the model performance. Therefore, for the sake of convenience, we may just utilize all of the non-color attributes of the source dataset. 

Since each non-color attribute is defined as a binary label, it has both positive and negative samples in the source dataset. In the domain adaptation process, the framework will assign an attribute all to be negative if positive samples of this attribute do not exist in the target dataset. For such reason, within the same category, any non-color attributes of the source dataset can be used in forming the shared-label space, although some attributes defined in the source data might all be negative in the target data.

\subsection{Trade-off Parameter Selection}
\begin{figure}[h!]
\centering 
\begin{tabular}{c}
\includegraphics[width=0.65\columnwidth]{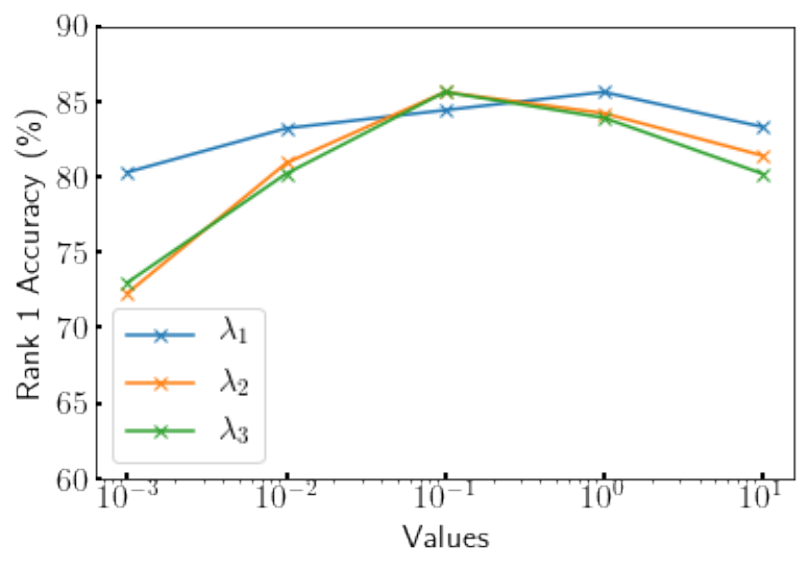} \\
\end{tabular}
\caption{Analysis on trade-off parameter selection.}
\label{fig:trade-off}
\end{figure}

In Fig. \ref{fig:trade-off}, we explore the sensitivity of trade-off parameters defined in Equation~\ref{eq:opt_goal}. From value $0.001$ to $10$, we vary one of them by fixing others as our default setting. It shows that $\lambda_{1}$ is not sensitive but very useful in IHDA framework. When $\lambda_{2}$ and $\lambda_{3}$ are assigned to $0.001$, it yields a poor retrieval performance compared with assigning them to $0.1$. This implies that attribute learning and domain adaptation are greatly beneficial to IHDA framework.

\tabcolsep=3pt

\tabcolsep=3pt
\begin{table*}[t]
\centering
\caption{The experimental settings and results of ablation studies. In the table, `Mark.' stands for the Market-1501 dataset, and `PKU.' stands for the PKU-Sketch dataset.}
\label{t:ablation}
\resizebox{\linewidth}{!}{
\begin{tabular}{c|l|C{7mm}C{7mm}C{7mm}C{7mm}C{7mm}C{7mm}|C{7mm}C{7mm}C{7mm}C{7mm}C{7mm}C{7mm}C{7mm}C{9mm}|c|cccc}
\toprule
\multirow{2}{*}{\textbf{Index}}&\multirow{2}{*}{\textbf{Description}} 
&\multicolumn{6}{c|}{\textbf{Modules}} 
&\multicolumn{8}{c|}{\textbf{Losses}}
&\multirow{2}{*}{\textbf{Training Data}} 
&\multirow{2}{*}{\textbf{R1}} 
&\multirow{2}{*}{\textbf{R5}} 
&\multirow{2}{*}{\textbf{R10}} 
&\multirow{2}{*}{\textbf{R20}} \\
&      
&$E_{1}$ &$E_{2}$ &$C^{s}_{id}$ &$C^{t1,t2}_{id}$ &$C_{att}$ &$C_{d}$ 
&$\mathcal{L}^{s}_{id}$ &$\mathcal{L}^{t1,t2}_{id}$ &$\mathcal{L}^{s}_{tri}$ &$\mathcal{L}^{t1,t2}_{tri}$ &$\mathcal{L}^{s}_{att}$ &$\mathcal{L}^{t1,t2}_{att}$ &$\mathcal{L}_{d}$ &$\mathcal{L}_{con}^{t1,t2}$
& 
& & & & \\
\midrule
(1)&w/o $\mathcal{X}^{s}$\&$\mathcal{X}^{t1,t2}$ &\checkmark  &\checkmark  & &  &   &  &  &  &  &  &  &  & &    &ImageNet
 &2.0        &8.0        &20.0      &40.0    \\ \midrule
(2)&w/o $\mathcal{X}^{t1,t2}$ &\checkmark  &\checkmark  &\checkmark &  & \checkmark&   &\checkmark & & \checkmark &     & \checkmark  &   &   &          &Mark.
 &8.0        &30.0       &44.0      &58.0    \\ \midrule
(3)&w/o $\mathcal{X}^{s}$ &\checkmark  &\checkmark  & & \checkmark    & &  & &\checkmark  & & \checkmark &      &   &    &    &PKU.(100\%) &32.0       &58.6      &68.2      &82.6    \\ \midrule                     
(4)&w/ 50\% $\mathcal{X}^{t1,t2}$    &\checkmark  &\checkmark  &\checkmark  &\checkmark  &\checkmark  &\checkmark  &\checkmark  &\checkmark &\checkmark &\checkmark &\checkmark &\checkmark &\checkmark &\checkmark   &Mark.+PKU.(50\%)
&75.4  &90.6 &92.0 &94.8 \\  \midrule                          
(5)&w/ 80\% $\mathcal{X}^{t1,t2}$   &\checkmark  &\checkmark  &\checkmark  &\checkmark  &\checkmark  &\checkmark  &\checkmark  &\checkmark &\checkmark &\checkmark &\checkmark &\checkmark &\checkmark  &\checkmark   &Mark.+PKU.(80\%)
&82.2  &92.6 &95.2 &98.0 \\  \midrule                            
(6)&w/o $\mathcal{L}_{att}^{s,t1,t2}$\&$\mathcal{L}_{d}$ &\checkmark  &\checkmark  &\checkmark  &\checkmark  &   &  &\checkmark  &\checkmark &\checkmark 
&\checkmark & & & &        &Mark.+PKU.(100\%)
&58.8       &78.0       &90.0       &94.2   \\ \midrule    
(7)&w/o $\mathcal{L}_{att}^{t1,t2}$\&$\mathcal{L}_{con}^{t1,t2}$\&$\mathcal{L}_{d}$ &\checkmark  &\checkmark  &\checkmark  &\checkmark  &\checkmark   &  &\checkmark  &\checkmark &\checkmark & {\checkmark}  & {\checkmark} & & &        &Mark.+PKU.(100\%)
&59.4       &79.2       &91.2       &95.8   \\ \midrule  
(8)& w/o $\mathcal{L}_{att}^{s,t1,t2}$ &\checkmark  &\checkmark  &\checkmark  &\checkmark  &    &\checkmark  &\checkmark  &\checkmark &\checkmark 
&\checkmark & & &\checkmark   &          &Mark.+PKU.(100\%)
&70.4       &82.0       &92.6       &96.2   \\ \midrule

 {(9)}& 
 {w/o $\mathcal{L}_{att}^{t1,t2}$\&$\mathcal{L}_{con}^{t1,t2}$} 
& {\checkmark}  & {\checkmark}   & {\checkmark}   & {\checkmark}     &  & {\checkmark}   
& {\checkmark}   & {\checkmark}   & {\checkmark}   & {\checkmark}     & {\checkmark}  & & {\checkmark}   &          & {Mark.+PKU.(100\%)}
& {70.6}       & {83.2}       & {92.4}       & {95.8}   \\ \midrule

(10)& w/o $\mathcal{L}_{con}^{t1,t2}$ &\checkmark  &\checkmark  &\checkmark  &\checkmark    & \checkmark  &\checkmark  &\checkmark  &\checkmark &\checkmark
&\checkmark &\checkmark &\checkmark  &\checkmark   &          &Mark.+PKU.(100\%)
&82.8       &94.2       &96.8       &98.4   \\ \midrule

 {(11)}&  {w/o $\mathcal{L}_{att}^{t1,t2}$} & {\checkmark}  & {\checkmark}  & {\checkmark}  & {\checkmark}    &  {\checkmark}  & {\checkmark}  & {\checkmark}  & {\checkmark} & {\checkmark}
& {\checkmark} & {\checkmark} & & {\checkmark}   &  {\checkmark}   
& {Mark.+PKU.(100\%)}
& {73.2}       & {85.6}       & {92.6}       & {96.8}   \\ \midrule

(12)& w/o $\mathcal{L}_{d}$  &\checkmark  &\checkmark  &\checkmark  &\checkmark   &\checkmark  &  &\checkmark  &\checkmark &\checkmark 
&\checkmark &\checkmark &\checkmark & & \checkmark &Mark.+PKU.(100\%)
&72.4       &84.2       &93.2       &96.6   \\ \midrule 

(13)&IHDA  &\checkmark  &\checkmark  &\checkmark  &\checkmark  &\checkmark  &\checkmark  &\checkmark  &\checkmark &\checkmark 
&\checkmark &\checkmark &\checkmark &\checkmark & \checkmark  &Mark.+PKU.(100\%)
&\textbf{85.6}  &\textbf{94.8} &\textbf{98.0} &\textbf{100.0} \\ 

\bottomrule
\end{tabular}
}
\end{table*}

\subsection{Ablation Studies}

\begin{figure*}[!ht]
\centering
  \includegraphics[width=0.95\linewidth]{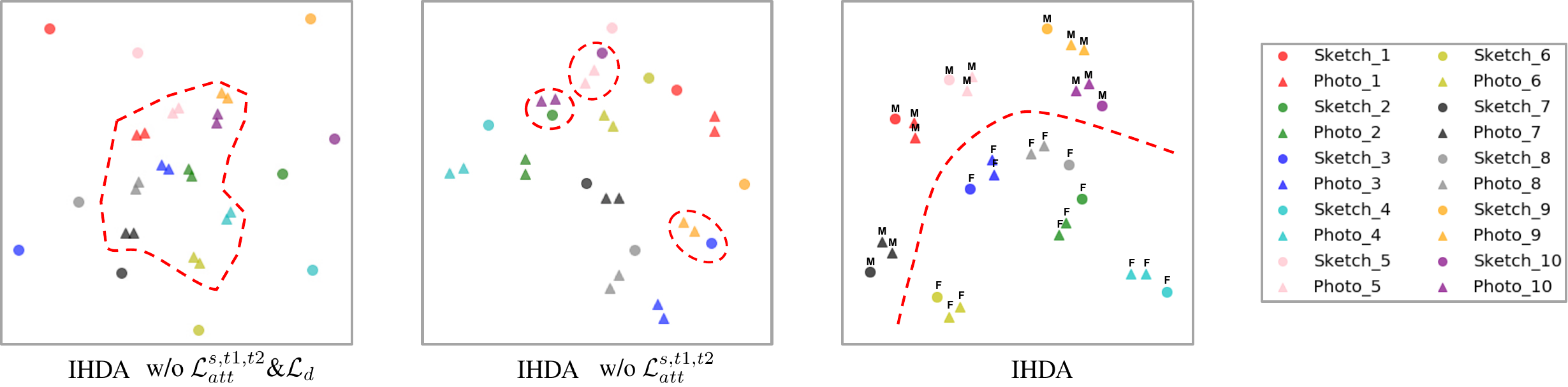}
  \caption{Visualization of embedding features for 10 identities in PKU-Sketch testing set by t-SNE~\cite{maaten2008visualizing}. Each color represents a specific identity by ground-truth annotation. Triangles and circles denote the photo and sketch embedding, respectively. Character ``M'' and ``F'' indicate predicted ``gender'' attribute as ``Male'' and ``Female'', resp. Red dot lines highlight where to look at. }
  \label{fig:embed}
\end{figure*}

\begin{figure}[!ht]
\centering
  \includegraphics[width=0.95\columnwidth]{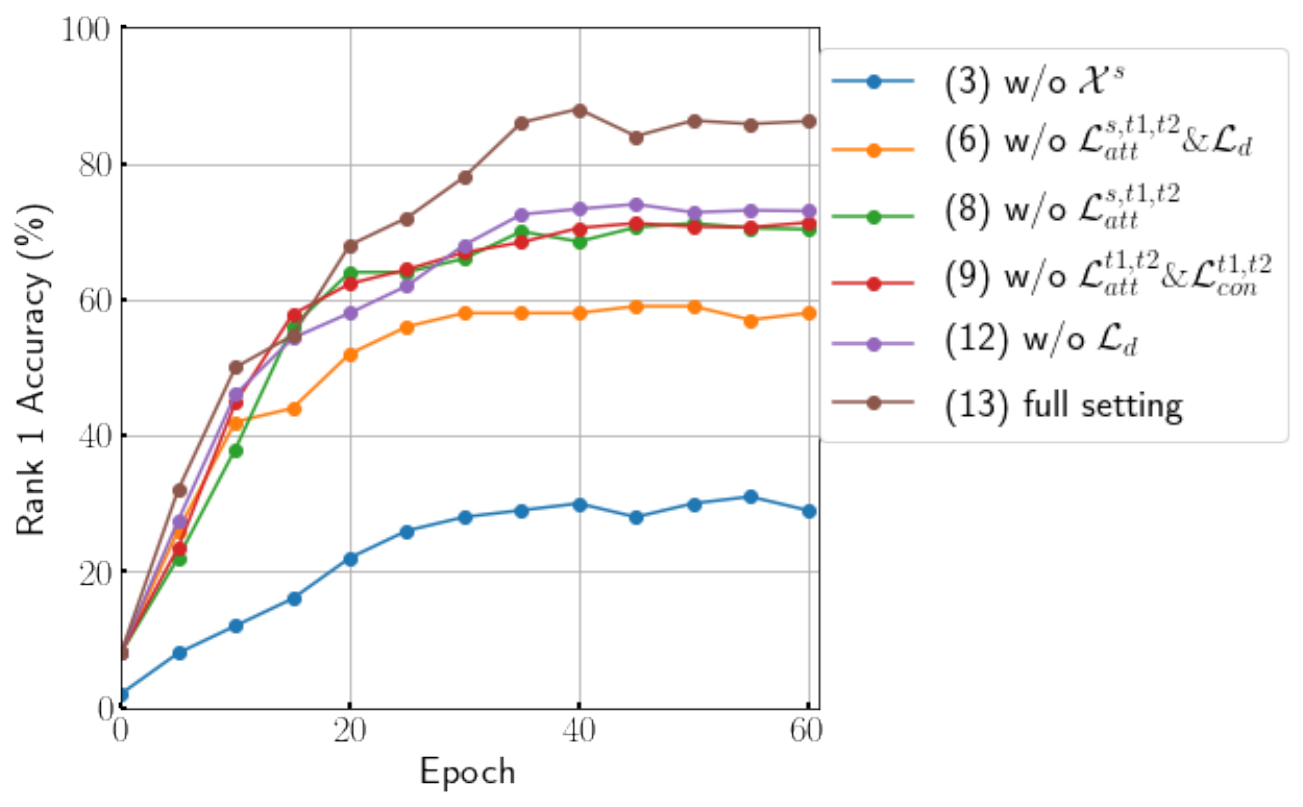}
  \caption{Rank 1 accuracy on PKU-Sketch target set as the number of training epochs goes up.}
  \label{fig:embed-2}
\end{figure}

To confirm the effectiveness of our IHDA framework, we conduct more detailed ablation studies by using PKU-Sketch and Market-1501 datasets. The experimental results are shown in TABLE \ref{t:ablation}. In TABLE \ref{t:ablation}, there are totally 11 variant settings for ablation studies. Among them, variant settings from (1) to (5) are used to investigate the IHDA performance at the different proportions of source and target data; variant settings from (6) to (12) are designed to study the effectiveness of each component in IHDA framework. We have the complete setup (i.e., components and datasets) of IHDA framework in (13), which considerably surpasses others.

Only using the pre-trained weights on ImageNet generates results that are almost identical to random guessing (see (1)). Due to the huge domain gaps between sketches and photos, solely using the pre-trained weights on Market-1501 (even with attributes) leads to poor performance (see (2)). Nevertheless, without external data, only training the model on PKU-Sketch also cannot obtain satisfactory results (see (3)) since lacking training data. Therefore, it is highly desirable to properly transfer the domain knowledge from Market-1501 to PKU-Sketch data. With complete components, the IHDA framework serves for such a purpose and generates promising results, even with $50\%$ and $80\%$ of PKU-Sketch training data (see (4) and (5)). 

It is critical to design the proper framework to transfer the domain knowledge from Market-1501 to PKU-Sketch data. Satisfactory retrieval accuracy cannot be approached through pre-training our model on Market-1501 data and then fine-tuning it on PKU-Sketch data (see (6)). In (7), although we apply attribute learning on source data, it does not significantly improve the target retrieval performance by a fine-tuning on target data. 

The alternative approach is to co-train the source and target data and reduce the domain gap between them. Without forming a shared label space via attributes, simply applying domain adaptation can roughly alleviate cross-domain discrepancies. Nonetheless, the instance-level source domain knowledge may not be properly transferred and leads to limited improvement (see (8)). On the other hand, without the domain adaptation, solely guided by shared attributes also cannot accomplish promising results (see (12)). These results are consistent with the analysis in Fig.~\ref{fig:ablation_att}. In (9), we show that target unsupervised attribute learning plays an important role in IHDA. The entropy-minimization loss and attribute-consistent loss work jointly to achieve optimal attribute-guided domain adaptation. Specifically, we verify the effectiveness of entropy-minimization and attribute-consistent in setting (10) and (11), respectively. The $\mathcal{L}^{t1,t2}_{att}$ contributes dominant effects while $\mathcal{L}^{t1,t2}_{con}$ can further improve the effectiveness.
 
Fig.~\ref{fig:embed-2} shows the Rank-1 accuracy on the PKU-sketch target set as the number of training epochs goes up. By using source data $\mathcal{X}$, The retrieval performance can increase faster than only using target data. Although there are many potential ways to utilize the source data, which includes aligned sketch-photo pairs and attributes, we propose the most effective one in the full setting proposal.

Furthermore, we analyze how embedding feature distributions change in different learning settings. For visualization purposes, we only select the gender attribute, as `Male' and `Female' in IHDA. By randomly picking up samples of 10 identities from the testing set of PKU-Sketch, we project their embedding features into a 2D space and show them in Fig.~\ref{fig:embed}. In simple fine-tuning strategy (i.e., TABLE \ref{t:ablation}.(6)), although both Market-1501 dataset and PKU-Sketch training set are utilized, photo embedding has a boundary to sketch embedding. By applying a DA without attribute guidance (i.e., TABLE \ref{t:ablation}.(7)), the boundary between photo and sketch is eliminated but the relative distance between some sketch-to-photo pairs could be incorrect. By using the complete IHDA, sketch-to-photo pairs have become more distinguishable with the attribute guidance.

\section{Discussion}
\label{sec:discusssion}

It is hard to judge ``pros'' or ``cons'' in our methods without considering the application scenario. 
Our IHDA framework is specifically designed for \textbf{instance-level heterogeneous image retrieval tasks}, it may not be applicable for category-level retrieval tasks. In instance-level retrieval tasks, the shared attributes are guaranteed since all instances belong to the same category. In contrast, it is challenging to find shared attributes for all instances in category-level retrieval tasks, such as Sketchy dataset~\cite{sangkloy2016sketchy}, PACS dataset~\cite{Li2017dg} and M3SDA dataset~\cite{peng2019moment},
Nonetheless, as the key contribution, our IHDA framework has overcome the limitations of applying simple fine-tuning strategy and conventional DA. It enables using rich-labeled photo retrieval data to remarkably improve the retrieval performance on limited-labeled sketch-photo retrieval tasks, which opens the door to train more effective cross-modal image retrieval models.

\section{Conclusion}
\label{sec:conclusion}

Nowadays, although a large amount of rich-labeled datasets have been made available to the public, how to utilize them to benefit a related limited-labeled task may vary case by case and is waiting for further exploration. Under such background, we analyze challenges for limited-labeled sketch-to-photo retrieval from the transfer learning perspective and raise an Instance-level Heterogeneous Domain Adaptation framework to tackle them. It demonstrates an important insight: attributes, which are only annotated in the source domain, can be used to form a shared label space for both source and target domains in instance-level retrieval task. With such a shared-label space, instance-level domain knowledge can be well transferred across datasets and heterogeneous modalities. It opens the door to train more effective cross-modal image retrieval models by using related rich-labeled single-modal image data.

\section*{Acknowledgment}
This work was supported by JSPS KAKENHI Grant Numbers JP17H06101, and a MSRA Collaborative Research 2019 Grant Awarded by Microsoft Research Asia.

\ifCLASSOPTIONcaptionsoff
  \newpage
\fi

\bibliographystyle{IEEEtran}
\bibliography{sample-bibliography}

\begin{thebibliography}{10}
\providecommand{\url}[1]{#1}
\csname url@samestyle\endcsname
\providecommand{\newblock}{\relax}
\providecommand{\bibinfo}[2]{#2}
\providecommand{\BIBentrySTDinterwordspacing}{\spaceskip=0pt\relax}
\providecommand{\BIBentryALTinterwordstretchfactor}{4}
\providecommand{\BIBentryALTinterwordspacing}{\spaceskip=\fontdimen2\font plus
\BIBentryALTinterwordstretchfactor\fontdimen3\font minus
  \fontdimen4\font\relax}
\providecommand{\BIBforeignlanguage}[2]{{%
\expandafter\ifx\csname l@#1\endcsname\relax
\typeout{** WARNING: IEEEtran.bst: No hyphenation pattern has been}%
\typeout{** loaded for the language `#1'. Using the pattern for}%
\typeout{** the default language instead.}%
\else
\language=\csname l@#1\endcsname
\fi
#2}}
\providecommand{\BIBdecl}{\relax}
\BIBdecl

\bibitem{pan2010survey}
S.~J. Pan and Q.~Yang, ``A survey on transfer learning,'' \emph{{IEEE} Trans.
  Knowl. Data Eng.}, vol.~22, no.~10, pp. 1345--1359, 2010.

\bibitem{pei2018multi}
Z.~Pei, Z.~Cao, M.~Long, and J.~Wang, ``Multi-adversarial domain adaptation,''
  in \emph{Proceedings of the AAAI Conference on Artificial Intelligence},
  2018, pp. 3934--3941.

\bibitem{rozantsev2018beyond}
A.~Rozantsev, M.~Salzmann, and P.~Fua, ``Beyond sharing weights for deep domain
  adaptation,'' \emph{{IEEE} Trans. Pattern Anal. Mach. Intell.}, vol.~41,
  no.~4, pp. 801--814, 2019.

\bibitem{cao2018partial}
Z.~Cao, L.~Ma, M.~Long, and J.~Wang, ``Partial adversarial domain adaptation,''
  in \emph{Proceedings of the European Conference on Computer Vision}, 2018,
  pp. 139--155.

\bibitem{zou2019consensus}
H.~Zou, Y.~Zhou, J.~Yang, H.~Liu, H.~P. Das, and C.~J. Spanos, ``Consensus
  adversarial domain adaptation,'' in \emph{Proceedings of the AAAI Conference
  on Artificial Intelligence}, 2019, pp. 5997--6004.

\bibitem{zou2018unsupervised}
Y.~Zou, Z.~Yu, B.~V. K.~V. Kumar, and J.~Wang, ``Unsupervised domain adaptation
  for semantic segmentation via class-balanced self-training,'' in
  \emph{Proceedings of the European Conference on Computer Vision}, 2018, pp.
  297--313.

\bibitem{pang2018cross}
L.~Pang, Y.~Wang, Y.-Z. Song, T.~Huang, and Y.~Tian, ``Cross-domain adversarial
  feature learning for sketch re-identification,'' in \emph{Proceedings of the
  ACM international Conference on Multimedia}, 2018, pp. 609--617.

\bibitem{ouyang2016forgetmenot}
S.~Ouyang, T.~M. Hospedales, Y.~Song, and X.~Li, ``Forgetmenot: Memory-aware
  forensic facial sketch matching,'' in \emph{Proceedings of the {IEEE}
  Conference on Computer Vision and Pattern Recognition}, 2016, pp. 5571--5579.

\bibitem{wang2015sketch}
S.~Wang, J.~Zhang, T.~X. Han, and Z.~Miao, ``Sketch-based image retrieval
  through hypothesis-driven object boundary selection with hlr descriptor,''
  \emph{IEEE Transactions on Multimedia}, vol.~17, no.~7, pp. 1045--1057, 2015.

\bibitem{yu2016sketch}
Q.~Yu, F.~Liu, Y.-Z. Song, T.~Xiang, T.~M. Hospedales, and C.-C. Loy, ``Sketch
  me that shoe,'' in \emph{Proceedings of the IEEE Conference on Computer
  Vision and Pattern Recognition}, 2016, pp. 799--807.

\bibitem{zhang2016sketch}
Y.~Zhang, X.~Qian, X.~Tan, J.~Han, and Y.~Tang, ``Sketch-based image retrieval
  by salient contour reinforcement,'' \emph{IEEE Transactions on Multimedia},
  vol.~18, no.~8, pp. 1604--1615, 2016.

\bibitem{bhattacharjee2018query}
S.~D. Bhattacharjee, J.~Yuan, Y.~Huang, J.~Meng, and L.~Duan, ``Query adaptive
  multiview object instance search and localization using sketches,''
  \emph{IEEE Transactions on Multimedia}, vol.~20, no.~10, pp. 2761--2773,
  2018.

\bibitem{choi2019sketchhelper}
J.~Choi, H.~Cho, J.~Song, and S.~M. Yoon, ``Sketchhelper: Real-time stroke
  guidance for freehand sketch retrieval,'' \emph{IEEE Transactions on
  Multimedia}, vol.~21, no.~8, pp. 2083--2092, 2019.

\bibitem{wang2020beyond}
\BIBentryALTinterwordspacing
Z.~Wang, Z.~Wang, Y.~Zheng, Y.~Wu, W.~Zeng, and S.~Satoh, ``Beyond
  intra-modality: A survey of heterogeneous person re-identification,'' in
  \emph{Proceedings of the Twenty-Ninth International Joint Conference on
  Artificial Intelligence}, 2020, pp. 4973--4980. [Online]. Available:
  \url{https://doi.org/10.24963/ijcai.2020/692}
\BIBentrySTDinterwordspacing

\bibitem{semjitter}
A.~Yu and K.~Grauman, ``Semantic jitter: Dense supervision for visual
  comparisons via synthetic images,'' in \emph{Proceedings of the IEEE
  International Conference on Computer Vision}, 2017, pp. 5570--5579.

\bibitem{liu2015faceattributes}
Z.~Liu, P.~Luo, X.~Wang, and X.~Tang, ``Deep learning face attributes in the
  wild,'' in \emph{Proceedings of the IEEE International Conference on Computer
  Vision}, 2015, pp. 3730--3738.

\bibitem{zheng2015scalable}
L.~Zheng, L.~Shen, L.~Tian, S.~Wang, J.~Wang, and Q.~Tian, ``Scalable person
  re-identification: {A} benchmark,'' in \emph{Proceedings of the {IEEE}
  International Conference on Computer Vision}, 2015, pp. 1116--1124.

\bibitem{wang2019incremental}
Z.~Wang, J.~Jiang, Y.~Yu, and S.~Satoh, ``Incremental re-identification by
  cross-direction and cross-ranking adaption,'' \emph{IEEE Transactions on
  Multimedia}, vol.~21, no.~9, pp. 2376--2386, 2019.

\bibitem{zeng2020illumination}
\BIBentryALTinterwordspacing
Z.~Zeng, Z.~Wang, Z.~Wang, Y.~Zheng, Y.-Y. Chuang, and S.~Satoh,
  ``Illumination-adaptive person re-identification,'' \emph{IEEE Transactions
  on Multimedia}, 2020. [Online]. Available:
  \url{http://dx.doi.org/10.1109/TMM.2020.2969782}
\BIBentrySTDinterwordspacing

\bibitem{wang2015multi}
Z.~Wang, R.~Hu, Y.~Yu, C.~Liang, and W.~Huang, ``Multi-level fusion for person
  re-identification with incomplete marks,'' in \emph{Proceedings of the 23rd
  ACM international conference on Multimedia}, 2015, pp. 1267--1270.

\bibitem{wang2018incremental}
Z.~Wang, X.~Bai, M.~Ye, and S.~Satoh, ``Incremental deep hidden attribute
  learning,'' in \emph{Proceedings of the 26th ACM international conference on
  Multimedia}, 2018, pp. 72--80.

\bibitem{song2016deep}
J.~Song, Y.~Song, T.~Xiang, T.~M. Hospedales, and X.~Ruan, ``Deep multi-task
  attribute-driven ranking for fine-grained sketch-based image retrieval,'' in
  \emph{Proceedings of the British Machine Vision Conference}, 2016, pp.
  132.1--132.11.

\bibitem{song2017deep}
J.~Song, Q.~Yu, Y.~Song, T.~Xiang, and T.~M. Hospedales, ``Deep
  spatial-semantic attention for fine-grained sketch-based image retrieval,''
  in \emph{Proceedings of the {IEEE} International Conference on Computer
  Vision}, 2017, pp. 5552--5561.

\bibitem{wu2018light}
X.~Wu, R.~He, Z.~Sun, and T.~Tan, ``A light {CNN} for deep face representation
  with noisy labels,'' \emph{{IEEE} Trans. Information Forensics and Security},
  vol.~13, no.~11, pp. 2884--2896, 2018.

\bibitem{wu2018coupled}
X.~Wu, L.~Song, R.~He, and T.~Tan, ``Coupled deep learning for heterogeneous
  face recognition,'' in \emph{Proceedings of the AAAI Conference on Artificial
  Intelligence}, 2018, pp. 1679--1686.

\bibitem{deng2019residual}
Z.~Deng, X.~Peng, and Y.~Qiao, ``Residual compensation networks for
  heterogeneous face recognition,'' in \emph{Proceedings of the AAAI Conference
  on Artificial Intelligence}, 2019, pp. 8239--8246.

\bibitem{saito2017asymmetric}
K.~Saito, Y.~Ushiku, and T.~Harada, ``Asymmetric tri-training for unsupervised
  domain adaptation,'' in \emph{Proceedings of the International Conference on
  Machine Learning}, 2017, pp. 2988--2997.

\bibitem{lin2019improving}
Y.~Lin, L.~Zheng, Z.~Zheng, Y.~Wu, Z.~Hu, C.~Yan, and Y.~Yang, ``Improving
  person re-identification by attribute and identity learning,'' \emph{Pattern
  Recognition}, vol.~95, pp. 151--161, 2019.

\bibitem{luo2017label}
Z.~Luo, Y.~Zou, J.~Hoffman, and L.~F. Fei-Fei, ``Label efficient learning of
  transferable representations acrosss domains and tasks,'' in
  \emph{Proceedings of the Advances in Neural Information Processing Systems},
  2017, pp. 165--177.

\bibitem{liu2018deep}
D.~Liu, N.~Wang, C.~Peng, J.~Li, and X.~Gao, ``Deep attribute guided
  representation for heterogeneous face recognition.'' in \emph{Proceedings of
  the International Joint Conference on Artificial Intelligence}, 2018, pp.
  835--841.

\bibitem{gebru2017fine}
T.~Gebru, J.~Hoffman, and L.~Fei-Fei, ``Fine-grained recognition in the wild: A
  multi-task domain adaptation approach,'' in \emph{Proceedings of the IEEE
  International Conference on Computer Vision}, 2017, pp. 1349--1358.

\bibitem{cui2018large}
Y.~Cui, Y.~Song, C.~Sun, A.~Howard, and S.~Belongie, ``Large scale fine-grained
  categorization and domain-specific transfer learning,'' in \emph{Proceedings
  of the IEEE Conference on Computer Vision and Pattern Recognition}, 2018, pp.
  4109--4118.

\bibitem{Lin2018multi}
S.~Lin, H.~Li, C.-T. Li, and A.~C. Kot, ``Multi-task mid-level feature
  alignment network for unsupervised cross-dataset person re-identification,''
  in \emph{Proceedings of the British Machine Vision Conference}, 2018, pp.
  19--32.

\bibitem{wang2018transferable}
J.~Wang, X.~Zhu, S.~Gong, and W.~Li, ``Transferable joint attribute-identity
  deep learning for unsupervised person re-identification,'' in
  \emph{Proceedings of the IEEE Conference on Computer Vision and Pattern
  Recognition}, 2018, pp. 2275--2284.

\bibitem{ge2020mutual}
\BIBentryALTinterwordspacing
Y.~Ge, D.~Chen, and H.~Li, ``Mutual mean-teaching: Pseudo label refinery for
  unsupervised domain adaptation on person re-identification,'' in
  \emph{International Conference on Learning Representations}, 2020. [Online].
  Available: \url{https://openreview.net/forum?id=rJlnOhVYPS}
\BIBentrySTDinterwordspacing

\bibitem{song2020unsupervised}
L.~Song, C.~Wang, L.~Zhang, B.~Du, Q.~Zhang, C.~Huang, and X.~Wang,
  ``Unsupervised domain adaptive re-identification: Theory and practice,''
  \emph{Pattern Recognition}, p. 107173, 2020.

\bibitem{lopez2017gradient}
D.~Lopez-Paz and M.~Ranzato, ``Gradient episodic memory for continual
  learning,'' in \emph{Advances in Neural Information Processing Systems},
  2017, pp. 6467--6476.

\bibitem{aljundi2018memory}
R.~Aljundi, F.~Babiloni, M.~Elhoseiny, M.~Rohrbach, and T.~Tuytelaars, ``Memory
  aware synapses: Learning what (not) to forget,'' in \emph{Proceedings of the
  European Conference on Computer Vision (ECCV)}, 2018, pp. 139--154.

\bibitem{long2013transfer}
M.~Long, J.~Wang, G.~Ding, J.~Sun, and P.~S. Yu, ``Transfer feature learning
  with joint distribution adaptation,'' in \emph{Proceedings of the IEEE
  International Conference on Computer Vision}, 2013, pp. 2200--2207.

\bibitem{saito2019semi}
K.~Saito, D.~Kim, S.~Sclaroff, T.~Darrell, and K.~Saenko, ``Semi-supervised
  domain adaptation via minimax entropy,'' in \emph{Proceedings of the IEEE
  International Conference on Computer Vision}, 2019, pp. 8050--8058.

\bibitem{grandvalet2005semi}
Y.~Grandvalet and Y.~Bengio, ``Semi-supervised learning by entropy
  minimization,'' in \emph{Proceedings of the Advances in Neural Information
  Processing Systems}, 2004, pp. 529--536.

\bibitem{he2016deep}
K.~He, X.~Zhang, S.~Ren, and J.~Sun, ``Deep residual learning for image
  recognition,'' in \emph{Proceedings of the IEEE conference on Computer Vision
  and Pattern Recognition}, 2016, pp. 770--778.

\bibitem{ganin2015unsupervised}
Y.~Ganin and V.~Lempitsky, ``Unsupervised domain adaptation by
  backpropagation,'' in \emph{Proceedings of the International Conference on
  International Conference on Machine Learning}, 2015, pp. 1180--1189.

\bibitem{chechik2010large}
G.~Chechik, V.~Sharma, U.~Shalit, and S.~Bengio, ``Large scale online learning
  of image similarity through ranking,'' \emph{Journal of Machine Learning
  Research}, vol.~11, no. Mar, pp. 1109--1135, 2010.

\bibitem{shu2019transferable}
Y.~Shu, Z.~Cao, M.~Long, and J.~Wang, ``Transferable curriculum for
  weakly-supervised domain adaptation,'' in \emph{Proceedings of the AAAI
  Conference on Artificial Intelligence}, 2019, pp. 4951--4958.

\bibitem{wang2018deep}
M.~Wang and W.~Deng, ``Deep visual domain adaptation: {A} survey,''
  \emph{Neurocomputing}, vol. 312, pp. 135--153, 2018.

\bibitem{dai2018cross}
P.~Dai, R.~Ji, H.~Wang, Q.~Wu, and Y.~Huang, ``Cross-modality person
  re-identification with generative adversarial training.'' in
  \emph{Proceedings of the International Joint Conference on Artificial
  Intelligence}, 2018, pp. 677--683.

\bibitem{long2017deep}
M.~Long, H.~Zhu, J.~Wang, and M.~I. Jordan, ``Deep transfer learning with joint
  adaptation networks,'' in \emph{Proceedings of the International Conference
  on Machine Learning}, 2017, pp. 2208--2217.

\bibitem{zhang2017joint}
J.~Zhang, W.~Li, and P.~Ogunbona, ``Jdeep face recognitionoint geometrical and
  statistical alignment for visual domain adaptation,'' in \emph{Proceedings of
  the IEEE Conference on Computer Vision and Pattern Recognition}, 2017, pp.
  5150--5158.

\bibitem{wang2017balanced}
J.~Wang, Y.~Chen, S.~Hao, W.~Feng, and Z.~Shen, ``Balanced distribution
  adaptation for transfer learning,'' in \emph{Proceedings of the IEEE
  International Conference on Data Mining}, 2017, pp. 1129--1134.

\bibitem{zhao2019learning}
H.~Zhao, R.~T. Des~Combes, K.~Zhang, and G.~Gordon, ``On learning invariant
  representations for domain adaptation,'' in \emph{Proceedings of the
  International Conference on Machine Learning}, 2019, pp. 7523--7532.

\bibitem{bhatt2012memetically}
H.~S. Bhatt, S.~Bharadwaj, R.~Singh, and M.~Vatsa, ``Memetically optimized
  {MCWLD} for matching sketches with digital face images,'' \emph{{IEEE} Trans.
  Information Forensics and Security}, vol.~7, no.~5, pp. 1522--1535, 2012.

\bibitem{wang2019learning}
Z.~Wang, J.~Jiang, Y.~Wu, M.~Ye, X.~Bai, and S.~Satoh, ``Learning sparse and
  identity-preserved hidden attributes for person re-identification,''
  \emph{IEEE Transactions on Image Processing}, vol.~29, no.~1, pp. 2013--2025,
  2019.

\bibitem{kingma2014adam}
D.~P. Kingma and J.~Ba, ``Adam: A method for stochastic optimization,''
  \emph{arXiv preprint arXiv:1412.6980}, 2014.

\bibitem{Luo_2019_Strong_TMM}
H.~{Luo}, W.~{Jiang}, Y.~{Gu}, F.~{Liu}, X.~{Liao}, S.~{Lai}, and J.~{Gu}, ``A
  strong baseline and batch normalization neck for deep person
  re-identification,'' \emph{IEEE Transactions on Multimedia}, pp. 1--1, 2019.

\bibitem{eitz2012sbsr}
M.~Eitz, R.~Richter, T.~Boubekeur, K.~Hildebrand, and M.~Alexa, ``Sketch-based
  shape retrieval,'' \emph{{ACM} Trans. Graph.}, vol.~31, no.~4, pp.
  31:1--31:10, 2012.

\bibitem{russakovsky2015imagenet}
O.~Russakovsky, J.~Deng, H.~Su, J.~Krause, S.~Satheesh, S.~Ma, Z.~Huang,
  A.~Karpathy, A.~Khosla, M.~Bernstein \emph{et~al.}, ``Imagenet large scale
  visual recognition challenge,'' \emph{International Journal of Computer
  Vision}, vol. 115, no.~3, pp. 211--252, 2015.

\bibitem{parkhi2015deep}
O.~M. Parkhi, A.~Vedaldi, and A.~Zisserman, ``Deep face recognition,'' in
  \emph{Proceedings of the British Machine Vision Conference}, vol.~41, 2015,
  pp. 1--12.

\bibitem{SangkloyBHH16}
P.~Sangkloy, N.~Burnell, C.~Ham, and J.~Hays, ``The sketchy database: learning
  to retrieve badly drawn bunnies,'' \emph{{ACM} Trans. Graph.}, vol.~35,
  no.~4, pp. 119:1--119:12, 2016.

\bibitem{sangkloy2016sketchy}
------, ``The sketchy database: learning to retrieve badly drawn bunnies,''
  \emph{ACM Transactions on Graphics (TOG)}, vol.~35, no.~4, pp. 1--12, 2016.

\bibitem{guyon2003introduction}
I.~Guyon and A.~Elisseeff, ``An introduction to variable and feature
  selection,'' \emph{Journal of machine learning research}, vol.~3, no. Mar,
  pp. 1157--1182, 2003.

\bibitem{maaten2008visualizing}
L.~v.~d. Maaten and G.~Hinton, ``Visualizing data using t-sne,'' \emph{Journal
  of Machine Learning Research}, vol.~9, no. Nov, pp. 2579--2605, 2008.

\bibitem{Li2017dg}
D.~Li, Y.~Yang, Y.~Song, and T.~M. Hospedales, ``Deeper, broader and artier
  domain generalization,'' in \emph{Proceedings of the International Conference
  on Computer Vision}, 2017, pp. 5543--5551.

\bibitem{peng2019moment}
X.~Peng, Q.~Bai, X.~Xia, Z.~Huang, K.~Saenko, and B.~Wang, ``Moment matching
  for multi-source domain adaptation,'' in \emph{Proceedings of the IEEE
  International Conference on Computer Vision}, 2019, pp. 1406--1415.

\end{thebibliography}

\end{document}